\documentclass[10pt,twocolumn,letterpaper]{article}

\usepackage{cvpr}
\usepackage{times}
\usepackage{epsfig}
\usepackage{graphicx}
\usepackage{amsmath}
\usepackage{amssymb}

\usepackage{xfrac}
\usepackage{multirow}
\usepackage{threeparttable}
\usepackage{colortbl}
\usepackage{float}
\usepackage{hyperref}
\hypersetup{colorlinks}
\usepackage{authblk}
\cvprfinalcopy % *** Uncomment this line for the final submission

\def\cvprPaperID{8460} % *** Enter the CVPR Paper ID here

\ifcvprfinal\fi

\newcommand{\N}{\mathcal{N}}

\newcommand{\x}{\mathbf{x}}
\newcommand{\y}{\mathbf{y}}
\newcommand{\f}{\mathbf{f}}
\renewcommand{\Re}{\mathbb{R}}
\renewcommand{\c}{\mathbf{c}}

\renewcommand{\l}{\mathbf{l}}

\newcommand{\p}{\mathbf{p}}

\renewcommand{\u}{\mathbf{u}}
\newcommand{\s}{\mathbf{s}}

\newcommand{\nop}[1]{}

\newcommand{\tabincell}[2]{\begin{tabular}{@{}#1@{}}#2\end{tabular}}

\begin{document}

\makeatletter
\renewcommand\AB@affilsepx{,\quad \protect\Affilfont}
\makeatother

%%%%%%%%% TITLE
\title{Superpixel Segmentation with Fully Convolutional Networks\vspace{-1em}}
\author[1]{Fengting Yang$^*$}
\author[1]{ Qian Sun$^\dagger$}
\author[2]{ Hailin Jin$^\ddagger$}
\author[1]{ Zihan Zhou $^\mathsection$}
\affil[1]{The Pennsylvania State University}
\affil[2]{Adobe Research \authorcr {\tt\small $^8$fuy34@psu.edu, $^\dagger$uestcqs@gmail.com, $^\ddagger$hljin@adobe.com, $^\mathsection$zzhou@ist.psu.edu}\vspace{-1.5em}}

% this author setting cannot be interpreted by arxiv
% \author{Fengting Yang  $\quad$  Qian Sun\\ %$^1$
% The Pennsylvania State University\\ %$^1$
% {\tt\small fuy34@psu.edu, uestcqs@gmail.com}
% % For a paper whose authors are all at the same institution,
% % omit the following lines up until the closing ``}''.
% % Additional authors and addresses can be added with ``\and'',
% % just like the second author.
% % To save space, use either the email address or home page, not both

% % \and
% % Qian Sun \\
% % The Pennsylvania State University\\
% % University Park\\
% % {\tt\small uestcqs@gmail.com}

% \and
% Hailin Jin\\
% Adobe Research\\
% {\tt\small hljin@adobe.com}

% \and
% Zihan Zhou\\
% The Pennsylvania State University\\
% {\tt\small zzhou@ist.psu.edu}
% }

\maketitle
\thispagestyle{empty}
%%%%%%%%% ABSTRACT
\begin{abstract}
In computer vision, superpixels have been widely used as an effective way to reduce the number of image primitives for subsequent processing. But only a few attempts have been made to incorporate them into deep neural networks. One main reason is that the standard convolution operation is defined on regular grids and becomes inefficient when applied to superpixels. Inspired by an initialization strategy commonly adopted by traditional superpixel algorithms, we present a novel method that employs a simple fully convolutional network to predict superpixels on a regular image grid. Experimental results on benchmark datasets show that our method achieves state-of-the-art superpixel segmentation performance while running at about 50fps. Based on the predicted superpixels, we further develop a downsampling/upsampling scheme for deep networks with the goal of generating high-resolution outputs for dense prediction tasks.  Specifically, we modify a popular network architecture for stereo matching to simultaneously predict superpixels and disparities. We show that improved disparity estimation accuracy can be obtained on public datasets. 
\end{abstract}

%%%%%%%%% BODY TEXT
\section{Introduction}

% Besides the computational efficiency, superpixels have gained particular popularity in 3D related tasks, such as depth prediction, stereo matching, and optical flow, as they provide a natural mean to integrate parametric models into the computation. For example, a slanted-plane model~\cite{BirchfieldT99}, which represents each image region (i.e., a superpixel) as a planar surface, is often adopted for those tasks~\cite{LiuSH14, YamaguchiHMU12, YamaguchiMU14}. By enforcing structural regularity within the image regions, these approaches have demonstrated strong capability in handling textureless and occluded regions, and in preserving geometric scene structures.

In recent years, deep neural networks (DNNs) have achieved great success in a wide range of computer vision applications. The advance of novel neural architecture design and training schemes, however, often comes a greater demand for computational resources in terms of both memory and time. Consider the stereo matching task as an example. It has been empirically shown that, compared to traditional 2D convolution, 3D convolution on a 4D volume (height$\times$width$\times$disparity$\times$feature channels)~\cite{KendallMDH17} can better capture context information and learn representations for each disparity level, resulting in superior disparity estimation results. But due to the extra feature dimension, 3D convolution is typically operating on spatial resolutions that are lower than the original input image size for the time and memory concern. For example, CSPN~\cite{abs-1810-02695}, the top-1 method on the KITTI 2015 benchmark, conducts 3D convolution at $1/4$ of the input size and uses bilinear interpolation to upsample the predicted disparity volume for final disparity regression. To handle high resolution images (\eg, $2000 \times 3000$), HSM~\cite{YangMHR19}, the top-1 method on the Middlebury-v3 benchmark, uses a multi-scale approach to compute disparity volume at $1/8$, $1/16$, and $1/32$ of the input size. Bilinear upsampling is again applied to generate disparity maps at the full resolution. In both cases, object boundaries and fine details are often not well preserved in final disparity maps due to the upsampling operation.

In computer vision, superpixels provide a compact representation of image data by grouping perceptually similar pixels together. As a way to effectively reduce the number of image primitives for subsequent processing, superpixels have been widely adopted in vision problems such as saliency detection~\cite{YangZLRY13}, object detection~\cite{ShuDS13}, tracking~\cite{WangLYY11}, and semantic segmentation~\cite{GouldRCEK08}.
However, superpixels are yet to be widely adopted in the DNNs for dimension
reduction. One main reason is that, the standard convolution operation in the convolutional neural networks (CNNs) is defined on a regular image grid. While a few attempts have been made to modify deep architectures to incorporate superpixels~\cite{HeLLHY15, GaddeJKKG16, KwakHH17, SuzukiAKA18}, performing convolution over an irregular superpixel grid remains challenging.

To overcome this difficulty, we propose a deep learning method to learn superpixels on the regular grid. Our key insight is that it is possible to associate each superpixel with a regular image grid cell, a strategy commonly used by traditional superpixel algorithms~\cite{LevinshteinSKFDS09, WangZGWZ13, BerghBRG15, AchantaSSLFS12, LiC15, LiuYY016, AchantaS17} as an initialization step (see Figure~\ref{fig:grid}). Consequently, we cast superpixel segmentation as a task that aims to find association scores between image pixels and regular grid cells, and use a fully convolutional network (FCN) to directly predict such scores. Note that recent work~\cite{JampaniSLYK18} also proposes an end-to-end trainable network for this task, but this method uses a deep network to extract pixel features, which are then fed to a soft K-means clustering module to generate superpixels.

\begin{figure*}[t]
\centering
\includegraphics[height =1.1in]{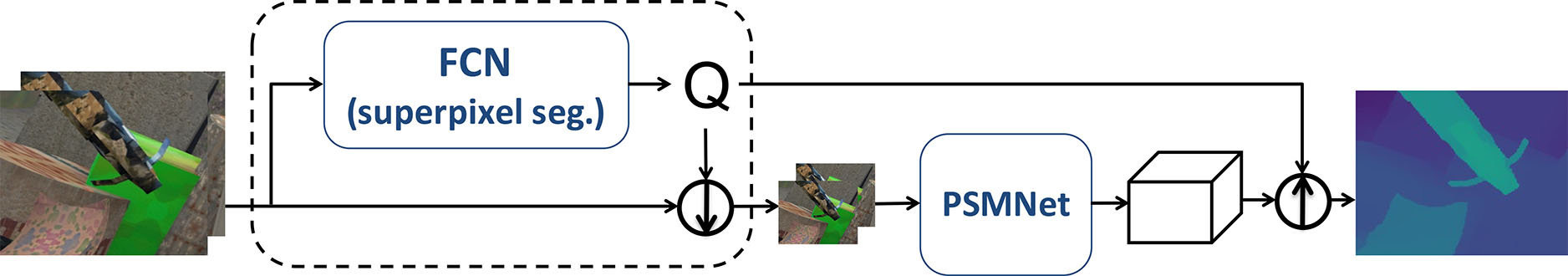} 
\caption{An illustration of our superpixel-based downsampling/upsampling scheme for deep networks. In this figure, we choose PSMNet~\cite{chang2018PSMNet} for stereo matching as our task network. The high-res input images are first downsampled using the superpixel association matrix $Q$ predicted by our superpixel segmentation network. To generate a high-res disparity map, we use the same matrix $Q$ to upsample the low-res disparity volume predicted by PSMNet for final disparity regression.}
\label{fig:framework}
\vspace{-4mm}
\end{figure*}

The key motivation for us to choose a standard FCN architecture is its simplicity as well as its ability to generate outputs on the regular grid. With the predicted superpixels, we further propose a general framework for downsampling/upsampling in DNNs. %deep neural networks.
As illustrated in Figure~\ref{fig:framework}, we replace the conventional operations for downsampling (\eg, stride-2 convolutions) and upsampling (\eg, bilinear upsampling) in the task network (PSMNet in the figure) with a superpixel-based downsampling/upsampling scheme to effectively preserve object boundaries and fine details. Further, the resulting network is end-to-end trainable.  One advantage of our joint learning framework is that superpixel segmentation is now directly influenced by the downstream task, and that the two tasks can naturally benefit from each other. In this paper, we take stereo matching as an example and show how the popular network PSMNet~\cite{chang2018PSMNet}, upon which many of the newest methods such as CSPN~\cite{abs-1810-02695} and HSM~\cite{YangMHR19} are built, can be adapted into our framework.

We have conducted extensive experiments to evaluate the proposed methods. For superpixel segmentation, experiment results on public benchmarks such as BSDS500~\cite{arbelaez2010bsds500} and NYUv2~\cite{Silberman2012nyuv2} demonstrate that our method is competitive with or better than the state-of-the-art w.r.t. a variety of metrics, and is also fast (running at about 50fps). For disparity estimation, our method outperforms the original PSMNet on SceneFlow~\cite{MayerIHFCDB16} as well as high-res datasets HR-VS~\cite{YangMHR19} and Middlebury-v3~\cite{scharstein2014middlebury}, verifying the benefit of incorporating superpixels into downstream vision tasks.

In summary, the {\bf main contributions} of the paper are: 
%\begin{itemize}
    %\item 
    1. We propose a simple fully convolutional network for superpixel segmentation, which achieves state-of-the-art performance on benchmark datasets.  %. Our method
    %\item 
    2. We introduce a general superpixel-based downsampling/upsampling framework for DNNs. We demonstrate improved accuracy in disparity estimation by incorporating superpixels into a popular stereo matching network. To the best of our knowledge, we are the first to develop a learning-based method that simultaneously perform superpixel segmentation and dense prediction.
%\end{itemize}  

\section{Related Work}
\noindent{\bf Superpixel segmentation.} There is a long line of research on superpixel segmentation, now a standard tool for many vision tasks. For a thorough survey on existing methods, we refer readers to the recent paper~\cite{StutzHL18}. Here we focus on methods which use a regular grid in the initialization step.

Turbopixels~\cite{LevinshteinSKFDS09} places initial seeds at regular intervals based on the desired number of superpixels, and grows them into regions until superpixels are formed. \cite{WangZGWZ13} grows the superpixels by clustering pixels using a geodesic distance that embeds structure and compactness constraints. SEEDS~\cite{BerghBRG15} initializes the superpixels on a grid, and continuously refines the boundaries by exchanging pixels between neighboring superpixels.

The SLIC algorithm~\cite{AchantaSSLFS12} employs K-means clustering to group nearby pixels into superpixels based on a 5-dimensional positional and CIELAB color features. Variants of SLIC include LSC~\cite{LiC15} which maps each pixel into a 10-dimensional feature space and performs weighted K-means, Manifold SLIC~\cite{LiuYY016} which maps the image to a 2-dimensional manifold to produce content-sensitive superpixels, and SNIC~\cite{AchantaS17} which replaces the iterative K-means clustering with a non-iterative region growing scheme. 

While the above methods rely on hand-crafted features, %and distance metrics, 
recent work~\cite{Tu0JSC0K18} proposes to learn pixel affinity from large data using DNNs. In~\cite{JampaniSLYK18}, the authors propose to learn pixel features which are then fed to a differentiable K-means clustering module for superpixel segmentation. The resulting method, SSN, is the first end-to-end trainable network for superpixel segmentation. Different from these methods, we train a deep neural network to directly predict the pixel-superpixel association map.

\smallskip
\noindent{\bf The use of superpixels in deep neural networks.} %deep neural networks (
Several methods propose to integrate superpixels into deep learning pipelines. These works typically use pre-computed superpixels %as a mean 
to manipulate learnt features so that important image properties (e.g., boundaries) can be better preserved. For example, \cite{HeLLHY15} uses superpixels to convert 2D image patterns into 1D sequential representations, which allows a DNN to efficiently explore long-range context for saliency detection. \cite{GaddeJKKG16} introduces a ``bilateral inception'' module which can be inserted into existing CNNs and perform bilateral filtering across superpixels, and \cite{KwakHH17, SuzukiAKA18} employ superpixels to pool features for semantic segmentation. Instead, we use superpixels as an effective way to downsample/upsample. Further, none of these works has attempted to jointly learn superpixels with the downstream tasks.

%\smallskip
%\noindent{\bf Deformable convolutional networks.} Besides superpixels, deformable convolution (DC) is another effective way to capture contextual information in DNNs. 
Besides, our method is also similar to the deformable convolutional network (DCN)~\cite{dai2017dcv,zhu2019dcv2} in that both can realize adaptive respective field. However, DCN is mainly designed to better handle geometric transformation and capture contextual information for feature extraction. Thus, unlike superpixels, a deformable convolution layer does not constrain that every pixel has to contribute to (thus is represented by) the output features.

 %However, DC is mainly designed to model geometric transformations according to the feature distribution, while our work is more focus on preserving object boundaries and fine details w.r.t. the image property.

\smallskip
\noindent{\bf Stereo matching.} Superpixel- or segmentation-based approach to stereo matching was first introduced in~\cite{BirchfieldT99}, and have since been widely used~\cite{HongC04, BleyerG04, KlausSK06, Wang08, BleyerRK10, GuneyG15}. These methods first segment the images into regions and fit a parametric model (typically a plane) to each region. In~\cite{YamaguchiHMU12, YamaguchiMU14}, Yamaguchi~\etal propose an optimization framework to jointly segments the reference image into superpixels and estimates the disparity map. \cite{LuoSU16} trains a CNN to predict initial pixel-wise disparities, which are refined using the slanted-plane MRF model. \cite{LeGendreBM17} develops an efficient algorithm which computes photoconsistency for only a random subset of pixels. Our work is fundamentally different from these optimization-based methods. Instead of fitting parametric models to the superpixels, we use superpixels to develop a new downsampling/upsampling scheme for DNNs. %deep neural networks.

In the past few years, deep networks~\cite{ZbontarL16, ShakedW17, PangSRYY17, YuWWJ18} taking advantage of large-scale annotated data have generated impressive stereo matching results. Recent methods~\cite{KendallMDH17, chang2018PSMNet, abs-1810-02695} employing 3D convolution achieve the state-of-the-art performance on public benchmarks. However, due to the memory constraints, these methods typically compute disparity volumes at a lower resolution. \cite{KhamisFRKVI18} bilinearly upsamples the disparity to the output size and refine it using an edge-preserving refinement network. Recent work~\cite{YangMHR19} has also explored efficient high-res processing, but its focus is on generating coarse-to-fine results to meet the need for anytime on-demand depth sensing in autonomous driving applications. 

\section{Superpixel Segmentation Method}

In this section, we introduce our CNN-based superpixel segmentation method. We first present our idea of directly predicting pixel-superpixel association on a regular grid in Section~\ref{sec:sp-grid}, followed by a description of our network design and loss functions in Section~\ref{sec:sp-method}. We further draw a connection between our superpixel learning regime with the recent convolutional spatial propagation (CSP) network~\cite{abs-1810-02695} for learning pixel affinity in Section~\ref{sec:sp-csp}. Finally, in Section~\ref{sec:sp-experiment}, we systematically evaluate our method on public benchmark datasets.

\subsection{Learning Superpixels on a Regular Grid}
\label{sec:sp-grid}

In the literature, a common strategy adopted~\cite{LevinshteinSKFDS09, WangZGWZ13, BerghBRG15, AchantaSSLFS12, LiC15, LiuYY016, AchantaS17, JampaniSLYK18} for superpixel segmentation is to first partition the $H\times W$ image using a regular grid of size $h\times w$ and consider each grid cell as an initial superpixel (i.e., a ``seed''). Then, the final superpixel segmentation is obtained by finding a mapping which assigns each pixel $\p=(u,v)$ to one of the seeds $\s = (i,j)$. Mathematically, we can write the mapping as $g_{\s}(\p) = g_{i,j}(u,v) = 1$ if the $(u,v)$-th pixel belongs to the $(i,j)$-th superpixel, and 0 otherwise.

\begin{figure}[t]
\centering
\includegraphics[height =1.3in]{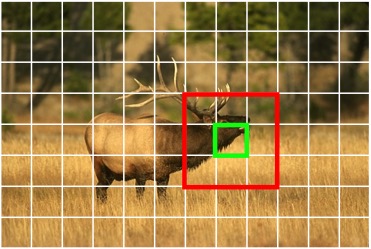} 
\caption{Illustration of $\N_{\p}$. For each pixel $\p$ in the green box, we consider the 9 grid cells in the red box for assignment.}\vspace{-3mm}
\label{fig:grid}
\end{figure}

In practice, however, it is unnecessary and computationally expensive to compute $g_{i,j}(u,v)$ for all pixel-superpixel pairs. Instead, for a given pixel $\p$, we constraint the search to the set of surrounding grid cells $\N_{\p}$. This is illustrated in Figure~\ref{fig:grid}. For each pixel $\p$ in the green box, we only consider the 9 grid cells in the red box for assignment. Consequently, we can write the mapping as a tensor $G \in \mathbb{Z}^{H\times W\times |\N_{\p}|}$ where $|\N_{\p}| = 9$.

\begin{figure*}[t]
\centering
\begin{tabular}{c|c}
%\hspace{-2mm}\includegraphics[height =0.41in]{figures/scheme1.pdf} &
\includegraphics[height =0.5in]{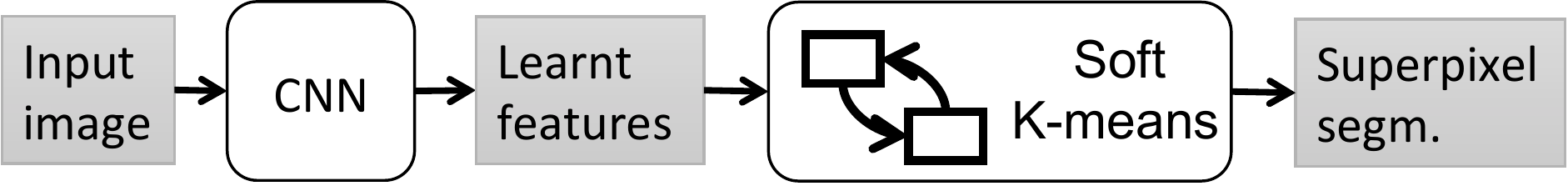} &
\includegraphics[height =0.5in]{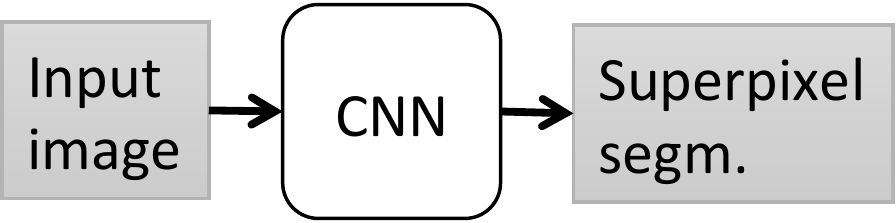} \\
(a) SSN & (b) Ours
\end{tabular}
\caption{Comparison of algorithmic schemes. SSN trains a CNN to extract pixel features, which are fed to an iterative K-means clustering module for superpixel segmentation. We train a CNN to directly generate superpixels by predicting a pixel-superpixel association map.}\vspace{-4mm}
\label{fig:comparison}
\end{figure*}

While several approaches~\cite{LevinshteinSKFDS09, WangZGWZ13, BerghBRG15, AchantaSSLFS12, LiC15, LiuYY016, AchantaS17, JampaniSLYK18} have been proposed to compute $G$, we take a different route in the paper. Specifically, we directly learn the mapping using a deep neural network. To make our objective function differentiable, we replace the hard assignment $G$ with a soft association map $Q\in \Re^{H\times W\times |\N_{\p}|}$. Here, the entry $q_{\s}(\p)$ represents the probability that a pixel $\p$ is assigned to each $\s\in \N_{\p}$, such that $\sum_{\s\in \N_{\p}} q_{\s}(\p) = 1$. Finally, the superpixels are obtained by assigning each pixel to the grid cell with the highest probability: $\s^* = \arg\max_{\s} q_{\s}(\p)$. 

Although it might seem a strong constraint that a pixel can only be associated to one of the 9 nearby cells, which leads to the difficulty to generate long/large superpixels, we want to emphasize the importance of the compactness. Superpixel is inherently an over-segmentation method. As one of the main purposes of our superpixel method is to perform the detail-preserved downsampling/upsampling to assist the downstream network, it is more important to capture spatial coherence in the local region. For the information goes beyond the 9-cell area, there is no problem to segment it into pieces and leave them for downstream network to aggregate with convolution operations. 

\smallskip
\noindent{\bf Our method vs. SSN~\cite{JampaniSLYK18}}. Recently, \cite{JampaniSLYK18} proposes SSN, an end-to-end trainable deep network for superpixel segmentation. Similar to our method, SSN also computes a soft association map $Q$. However, unlike our method, SSN uses the CNN as a means to extract pixel features, which are then fed to a soft K-means clustering module to compute $Q$.

We illustrate the algorithmic schemes of the two methods in Figure~\ref{fig:comparison}. Both SSN and our method can take advantage of CNN to learn complex features using task-specific loss functions. But unlike SSN, we combine feature extraction and superpixel segmentation into a single step. As a result, our network runs faster and can be easily integrated into existing CNN frameworks for downstream tasks (Section~\ref{sec:disparity}).

\subsection{Network Design and Loss Functions} 
\label{sec:sp-method}
 
\begin{figure}[t]
\centering
\includegraphics[width =2.8in]{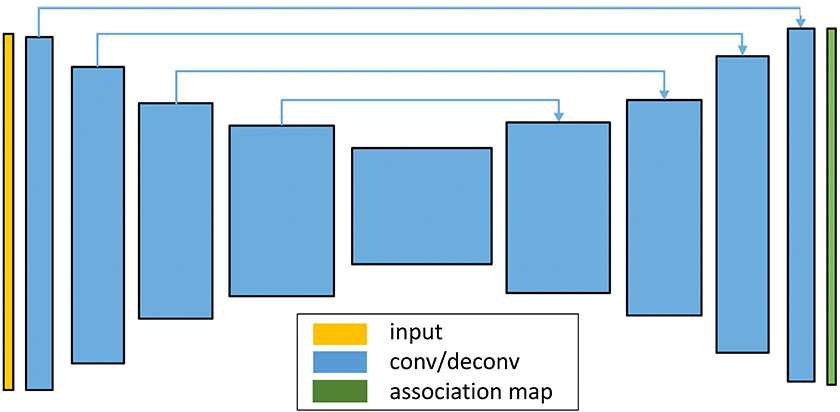} 
\caption{Our simple encoder-decoder architecture for superpixel segmentation. % (See supplementary for more details). 
%The yellow block indicates the input image, and the green block indicates the predicted association map $Q$. Each blue block represent two continuous convolutional layers or one deconvlution layer plus one convolutional layer, whose width and height of the block indicates the channel and the spatial dimension of the feature map.
Please refer to the supplementary materials for detailed specifications.
}\vspace{-3mm}
\label{fig:sp-net}
\end{figure}

As shown in Figure~\ref{fig:sp-net}, we use a standard encoder-decoder design with skip connections to predict superpixel association map $Q$. The encoder takes a color image as input and produces high-level feature maps via a convolutional network. The decoder then gradually upsamples the feature maps via deconvolutional layers to make final prediction, taking into account also the features from corresponding encoder layers. We use leaky ReLU for all layers except for the prediction layer, where softmax is applied. 
%Detailed specification of our network architecture is presented in the supplementary material. 

Similar to SSN~\cite{JampaniSLYK18}, one of the main advantages of our end-to-end trainable superpixel network is its flexibility w.r.t. the loss functions. Recall that the idea of superpixels is to group similar pixels together. For different applications, one may wish to define similarity in different ways. 
%For example, we may want to group pixels with the same color, semantic label, or other pixel properties. 

Generally, let $\f(\p)$ be the pixel property we want the superpixels to preserve. Examples of $\f(\p)$ include a 3-dimensional CIELAB color vector, and/or a $N$-dimensional one-hot encoding vector of semantic labels, where $N$ is the number of classes, and many others. We further represent a pixel's position by its image coordinates $\p = [x, y]^T$.

Given the predicted association map $Q$, we can compute the center of any superpixel $\s$, $\c_{\s} = (\u_{\s}, \l_{\s})$ where $\u_{\s}$ is the property vector and $\l_{\s}$ is the location vector, as follows: 
\begin{equation}
\u_{\s} = \frac{\sum_{\p: \s\in \N_{\p}} \f(\p) \cdot q_{\s}(\p)}{\sum_{\p: \s\in \N_{\p}} q_{\s}(\p)}, \quad 
\l_{\s} = \frac{\sum_{\p: \s\in \N_{\p}} \p \cdot q_{\s}(\p)}{\sum_{\p: \s\in \N_{\p}} q_{\s}(\p)}.
\label{eqn:center}
\end{equation}
Here, recall that $\N_{\p}$ is the set of surrounding superpixels of $\p$, and $q_{\s}(\p)$ is the network predicted probability of $\p$ being associated with superpixel $\s$. In Eq~\eqref{eqn:center}, each sum is taken over all the pixels with a possibility to be assigned to $\s$. 

Then, the reconstructed property and location of any pixel $\p$ are given by:
\begin{equation}
\f'(\p) = \sum_{\s\in \N_{\p}} \u_{\s} \cdot q_{\s}(\p), \quad \p' = \sum_{\s\in \N_{\p}} \l_{\s} \cdot q_{\s}(\p).
\label{eqn:pixel}
\end{equation}

Finally, the general formulation of our loss function has two terms. The first term encourages the trained model to group pixels with similar property of interest, and the second term enforces the superpixels to be spatially compact:
\begin{equation}
L(Q) = \sum_{\p} \textup{dist}(\f(\p), \f'(\p)) + \frac{m}{S}\|\p - \p'\|_2,
\label{eqn:loss}
\end{equation}
where $\textup{dist}(\cdot, \cdot)$ is the task specific distance metric depending on the pixel property $\f(\p)$, $S$ is the superpixel sampling interval, and $m$ is a weight balancing the two terms.  %importance of the

% Note that, to make the loss differentiable, in Eq.~\eqref{eqn:center} and~\eqref{eqn:pixel} we use soft association $Q$, instead of the hard assignment in the original SLIC algorithm.

In this paper, we consider two different choices of $\f(\p)$. First, we choose the CIELAB color vector and use the $\ell_2$ norm as the distance measure. This leads to an objective function similar to the original SLIC method~\cite{AchantaSSLFS12}: 
\begin{equation}
L_{SLIC}(Q) = \sum_{\p} \|\f_{col}(\p) - \f'_{col}(\p)\|_2 + \frac{m}{S}\|\p - \p'\|_2.
\label{eqn:loss-slic}
\end{equation}

Second, following~\cite{JampaniSLYK18}, we choose the one-hot encoding vector of semantic labels and use cross-entropy $E(\cdot,\cdot)$ as the distance measure: %\todo we may need to unify the loss here
\begin{equation}
L_{sem}(Q) = \sum_{\p} E(\f_{sem}(\p), \f'_{sem}(\p)) + \frac{m}{S}\|\p - \p'\|_2.
\label{eqn:loss-sem}
\end{equation}

\subsection{Connection to Spatial Propagation Network}
\label{sec:sp-csp}

Recently, \cite{abs-1810-02695} proposes the convolutional spatial propagation (CSP) network, which learns an affinity matrix to propagate information to nearly spatial locations. By integrating the CSP module into existing deep neural networks, \cite{abs-1810-02695} has demonstrated improved performance in affinity-based vision tasks such as depth completion and refinement. In this section, we show that the computation of superpixel centers using learnt association map $Q$ can be written mathematically in the form of CSP, thus draw a connection between learning $Q$ and learning the affinity matrix as in~\cite{abs-1810-02695}.

Given an input feature volume $X\in\Re^{H\times W\times C}$, the convolutional spatial propagation (CSP) with a kernel size $K$ and stride $S$ can be written as:
\begin{equation}
\y_{i,j} = \sum_{a,b=-K/2+1}^{K/2} \kappa_{i,j}(a,b) \odot \x_{i\cdot S + a, j\cdot S + b},
\label{eq:aff}
\end{equation}
where $Y\in\Re^{h \times w \times C}$ is an output volume such that $h=\frac{H}{S}$ and $w = \frac{W}{S}$, $\kappa_{i,j}$ is the output from an affinity network such that $\sum_{a,b=-K/2+1}^{K/2}\kappa_{i,j}(a,b) = 1$, and $\odot$ is element-wise product.

In the meantime, as illustrated in Figure~\ref{fig:grid}, to compute the superpixel center associated with the $(i,j)$-th grid cell, we consider all pixels in the surrounding $3S\times 3S$ region:
\begin{equation}
\c_{i,j} = \sum_{a,b=-3S/2+1}^{3S/2} \hat{q}_{i,j}(a,b) \odot \x_{i\cdot S + a, j\cdot S + b},
\label{eq:center}
\end{equation}
where 
\begin{equation}
\setlength{\abovedisplayskip}{-2pt}
\hat{q}_{i,j}(a,b) = \frac{q_{i,j}(u,v)}{\sum_{a,b=-3S/2+1}^{3S/2} q_{i,j}(u,v)},
\end{equation}
and $u=i\cdot S + a, v=j\cdot S + b$.

Comparing Eq.~\eqref{eq:aff} with Eq.~\eqref{eq:center}, we can see that computing the center of superpixel of size $S\times S$ is equivalent to performing CSP with a $3S\times 3S$ kernel derived from $Q$. Furthermore, both $\kappa_{i,j}(a,b)$ and $q_{i,j}(u,v)$ represent the learnt weight between the spatial location $(u,v)$ in the input volume and $(i,j)$ in the output volume. In this regard, predicting $Q$ in our work can be viewed as learning an affinity matrix as in~\cite{abs-1810-02695}.

Nevertheless, we point out that, while the techniques presented in this work and \cite{abs-1810-02695} share the same mathematical form, they are developed for very different purposes. In~\cite{abs-1810-02695}, Eq.~\eqref{eq:aff} is employed repeatedly (with $S=1$) to propagate information to nearby locations, whereas in this work, we use Eq.~\eqref{eq:center} to compute superpixel centers (with $S>1$).

\begin{figure*}[t]
\centering
\begin{tabular}{ccc}
%\hspace{-2mm}\includegraphics[height =0.41in]{figures/scheme1.pdf} &
\includegraphics[height =1.6in]{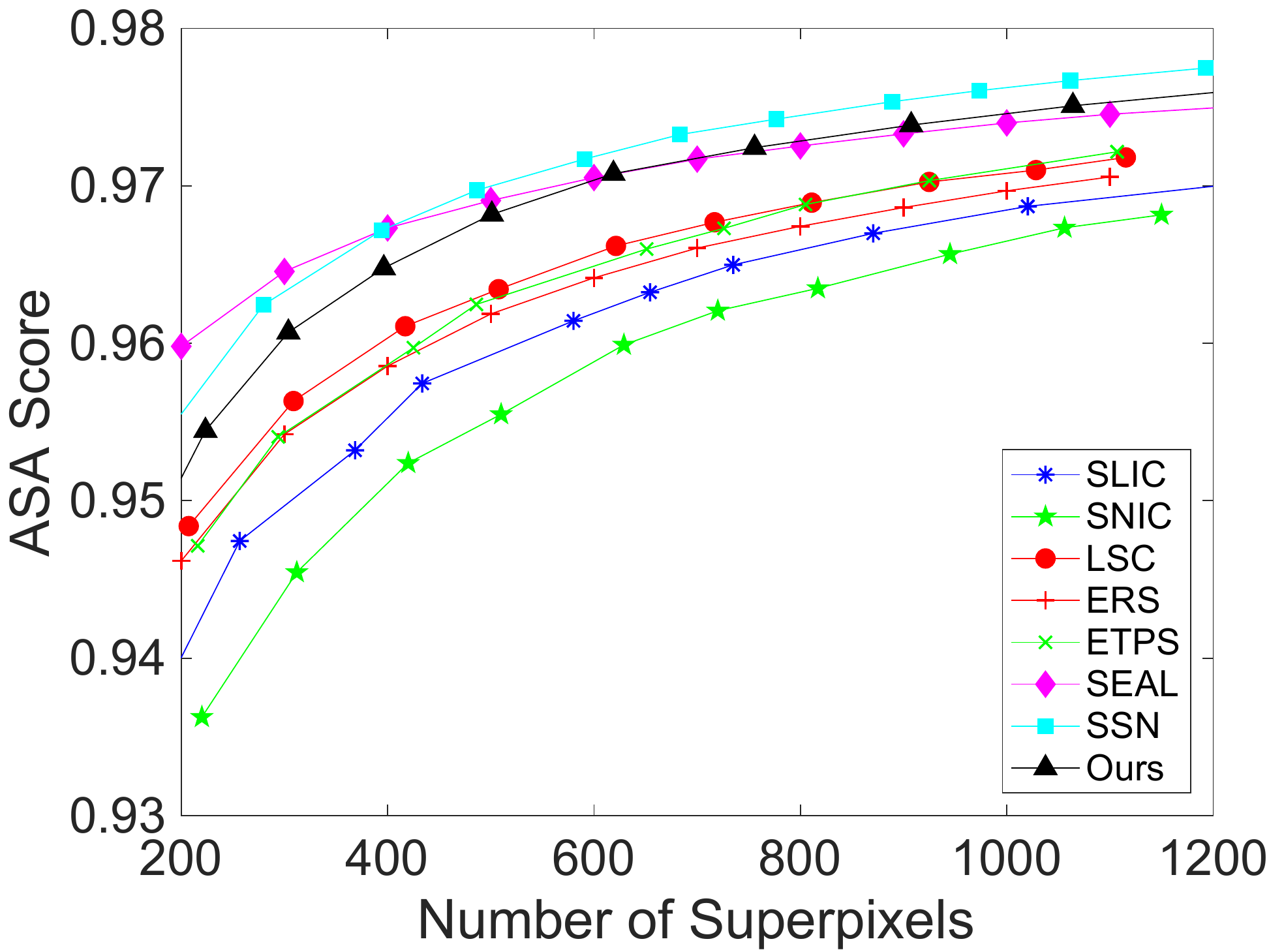} &
\includegraphics[height =1.6in]{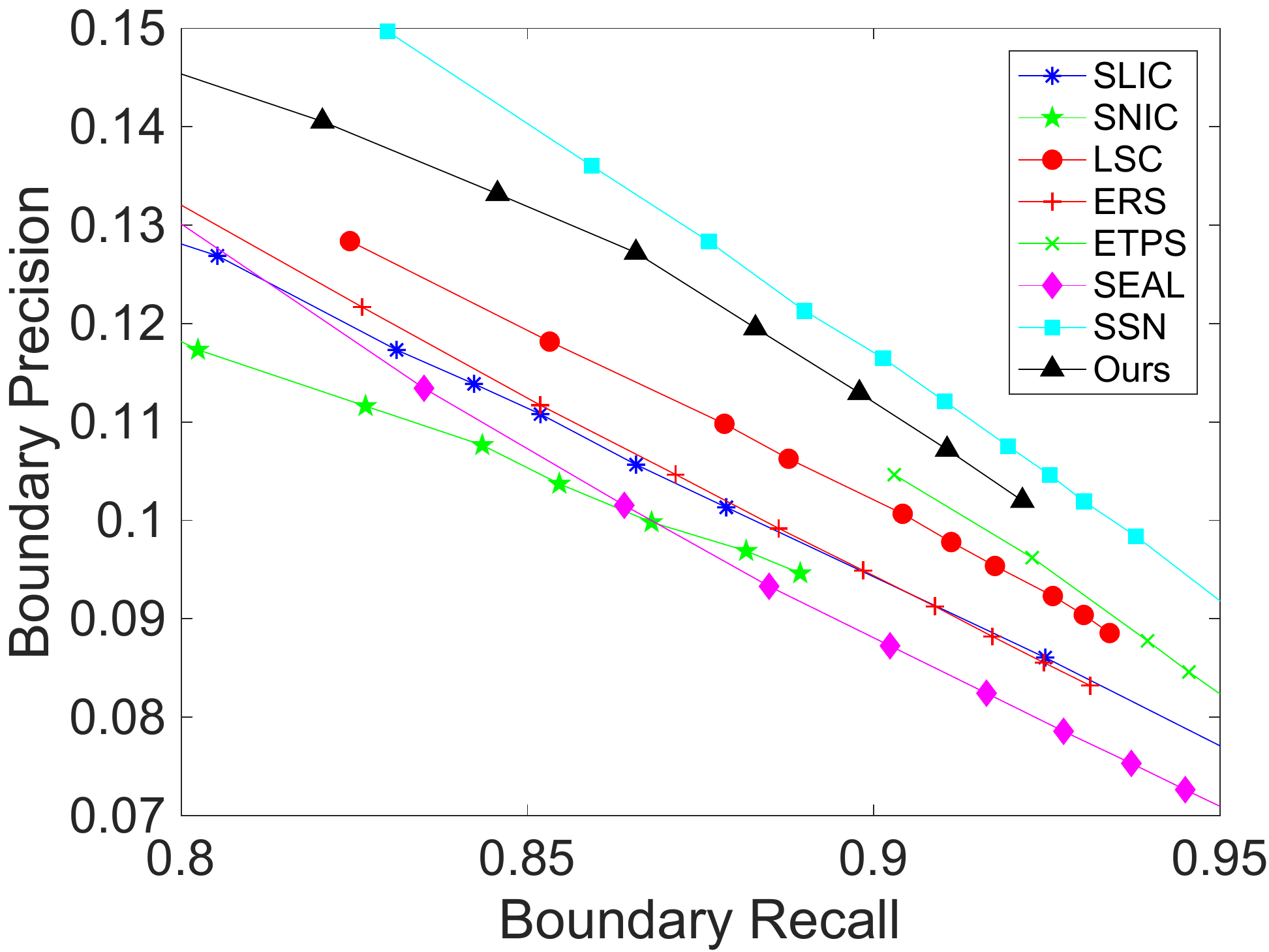} &
\includegraphics[height =1.6in]{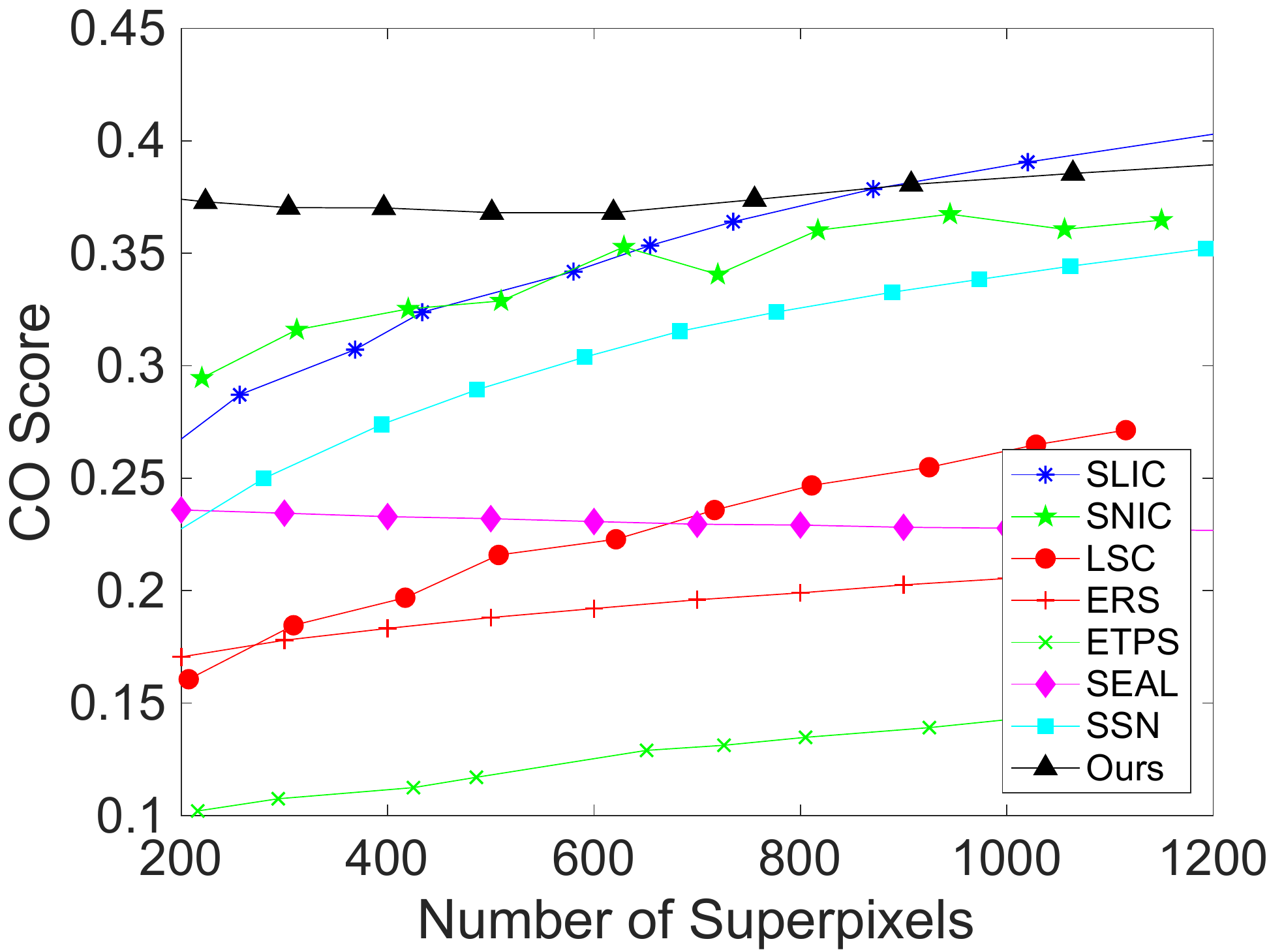}
\end{tabular}
\caption{Superpixel segmentation results on BSDS500. {\bf From left to right}: ASA, BR-BP, and CO.}\vspace{-3mm}
\label{fig:BSDS_res}
\end{figure*}

\begin{figure*}[t]
\centering
\begin{tabular}{ccc}
%\hspace{-2mm}\includegraphics[height =0.41in]{figures/scheme1.pdf} &
\includegraphics[height =1.6in]{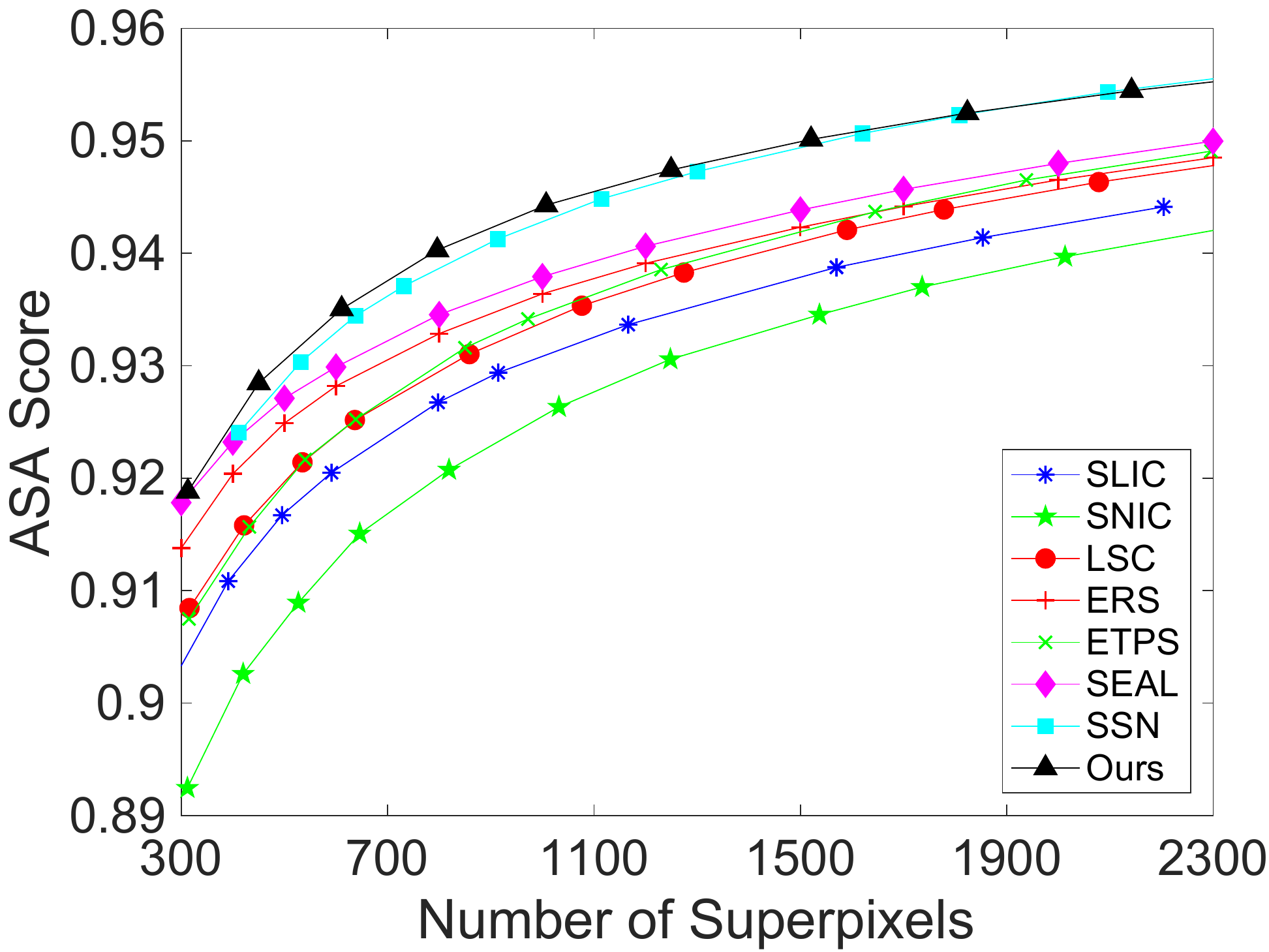} &
\includegraphics[height =1.6in]{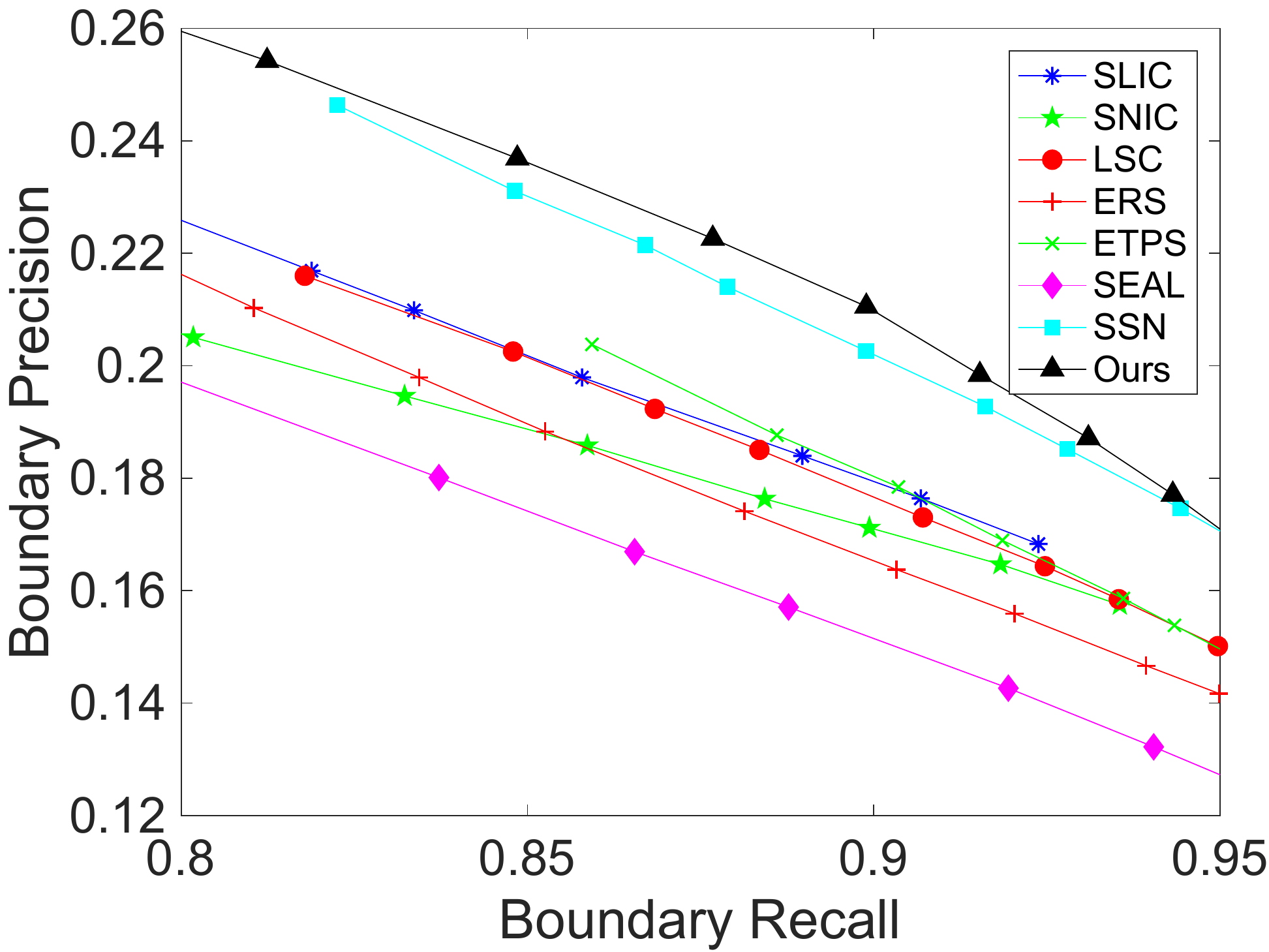} &
\includegraphics[height =1.6in]{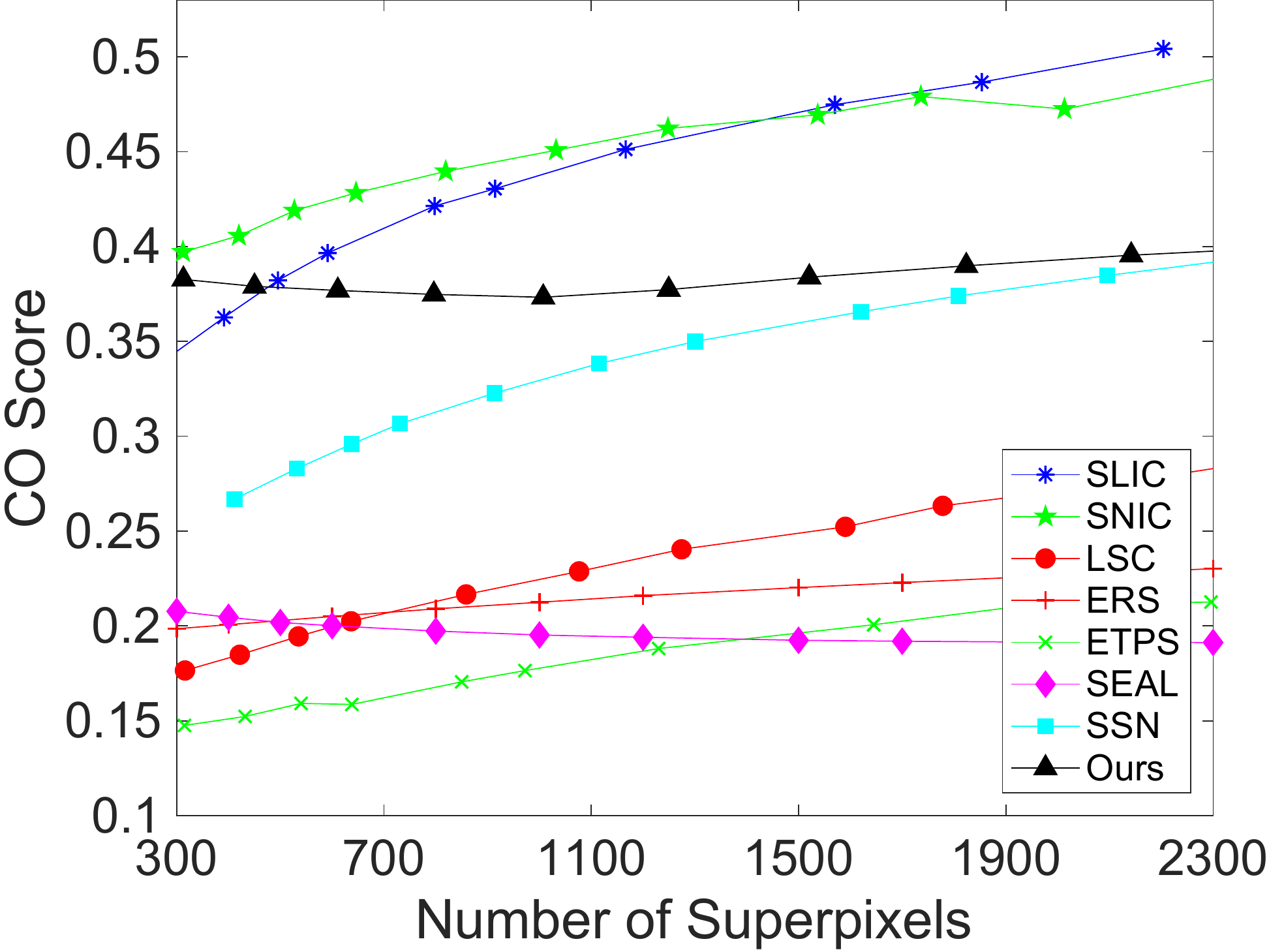}
\end{tabular}
\caption{Superpixel segmentation results on NYUv2. {\bf From left to right}: ASA, BR-BP, and CO.}\vspace{-4mm}
\label{fig:NYU_res}
\end{figure*}

\subsection{Experiments}
\label{sec:sp-experiment}

We train our model with segmentation labels on the standard benchmark BSDS500~\cite{arbelaez2010bsds500} and compare it with state-of-the-art superpixel methods. To further evaluate the method generalizability, we also report its performances without fine-tuning on another benchmark dataset NYUv2~\cite{Silberman2012nyuv2}.

%We train our network with segmentation labels on the standard benchmark dataset BSDS500~\cite{arbelaez2010bsds500} and compare its performance with state-of-the-art superpixel methods. To further evaluate the generalizability of our method, we also report the performance of the trained model without fine-tuning on another benchmark dataset NYUv2~\cite{Silberman2012nyuv2}.

All evaluations are conducted using the protocols and codes provided by~\cite{StutzHL18}\footnote{\url{https://github.com/davidstutz/superpixel-benchmark}}.
We run LSC~\cite{LiC15}, ERS~\cite{liu2011entropy}, SNIC~\cite{AchantaS17}, SEAL~\cite{Tu0JSC0K18}, and SSN~\cite{JampaniSLYK18} with the original implementations from the authors, and run SLIC~\cite{AchantaSSLFS12} and ETPS~\cite{yao2015etps}  with the codes provided in~\cite{StutzHL18}. For LSC, ERS, SLIC and ETPS, we use the best parameters reported in \cite{StutzHL18}, and for the rest, we use the default parameters recommended by the original authors. 

% ===== time
\begin{figure}[ht]
\centering
\includegraphics[height =1.6in]{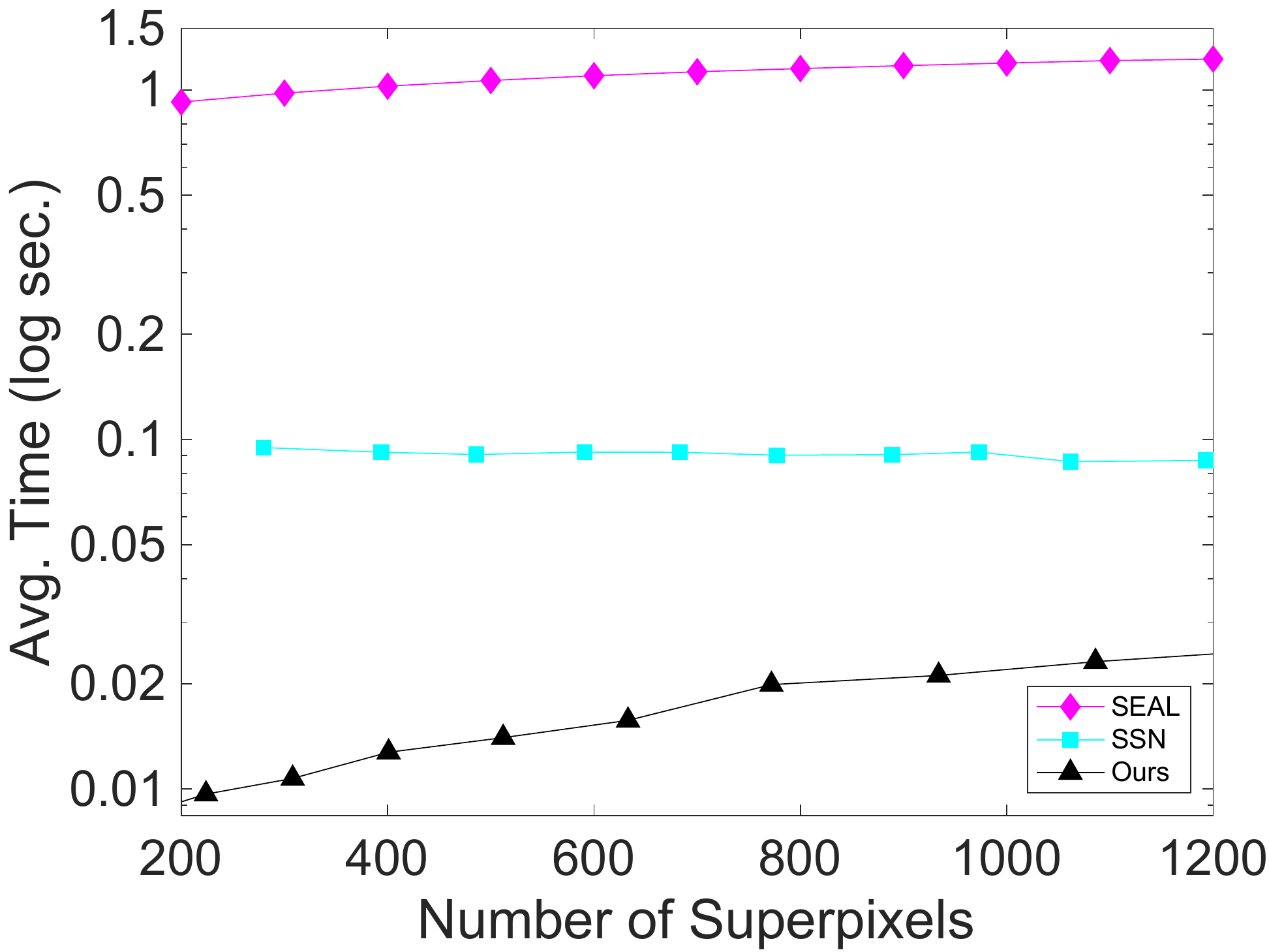} 
\caption{Average runtime of different DL methods w.r.t. number of superpixels. Note that $y$-axis is plotted in the logarithmic scale.}\vspace{-4mm}
\label{fig:time}
\end{figure}
% === 

\smallskip
\noindent{\bf Implementation details.} Our model is implemented with PyTorch, and optimized using Adam with $\beta_1=0.9$ and $\beta_2=0.999$. We use $L_{sem}$ in Eq.~\eqref{eqn:loss-sem} for this experiment, with $m=0.003$. During training, we randomly crop the images to size $208 \times 208$ as input, and perform horizontal/vertical flipping for data augmentation. The initial learning rate is set to $5 \times 10^{-5}$, and is reduced by half after 200k iterations. Convergence is reached at about 300k iterations. 
%All the experiments in the paper are performed on a Nvidia GTX 1080 Ti GPU device. 

For training, we use a grid with cell size $16\times 16$, which is equivalent to setting the desired number of superpixels to 169. At the test time, to generate varying number of superpixels, we simply resize the input image to the appropriate size. For example, by resizing the image to $480\times 320$, our network will generate about 600 superpixels. Furthermore, for fair comparison, most evaluation protocols expect superpixels to be spatially connected. To enforce that, we apply an off-the-shelf component connection algorithm to our output, which merges superpixels that are smaller than a certain threshold with the surrounding ones.\footnote{Code and models are available at \url{https://github.com/fuy34/superpixel_fcn}.}

\begin{figure*}[t]
\centering
% \begin{tabular}{cccccc}
% \hspace{-3.5mm}\includegraphics[height =0.77in]{Input} & GT segments & SLIC & SEAL & SSN & Ours
% \end{tabular}
\begin{tabular}{cccccc}
%\hspace{-2mm}\includegraphics[height =0.41in]{figures/scheme1.pdf} &
\hspace{-3.5mm}Input & \hspace{-3.5mm}GT segments & \hspace{-3.5mm}SLIC & \hspace{-3.5mm}SEAL & \hspace{-3.5mm}SSN & \hspace{-3.5mm}Ours\\
\hspace{-3.5mm}\includegraphics[height =0.76in]{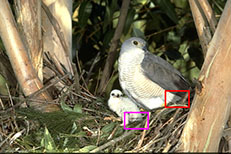} &
\hspace{-3.5mm}\includegraphics[height= 0.76in]{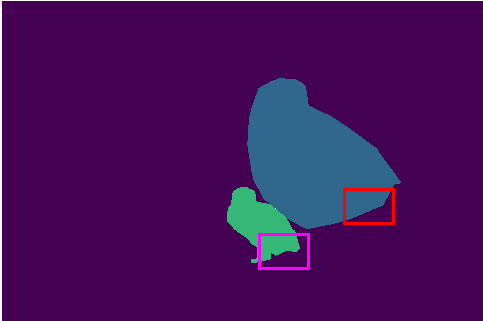} &
\hspace{-3.5mm}\includegraphics[height =0.76in]{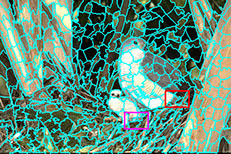} &
\hspace{-3.5mm}\includegraphics[height =0.76in]{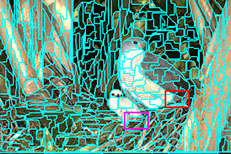} &
\hspace{-3.5mm}\includegraphics[height =0.76in]{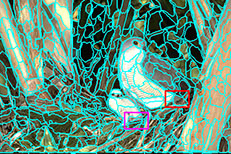} &
\hspace{-3.5mm}\includegraphics[height =0.76in]{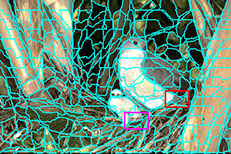} 
\end{tabular}
\begin{tabular}{cccccccccccc}
%\hspace{-2mm}\includegraphics[height =0.41in]{figures/scheme1.pdf} &
\hspace{-3.3mm}\includegraphics[height =0.40in]{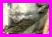} &
\hspace{-4.0mm}\includegraphics[height =0.40in]{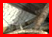} &
\hspace{-3.3mm}\includegraphics[height= 0.40in]{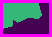} &
\hspace{-4.0mm}\includegraphics[height= 0.40in]{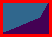} &
\hspace{-3.3mm}\includegraphics[height =0.40in]{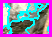} &
\hspace{-4.0mm}\includegraphics[height =0.40in]{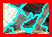} &
\hspace{-3.3mm}\includegraphics[height =0.40in]{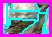} &
\hspace{-4.0mm}\includegraphics[height =0.40in]{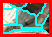} &
\hspace{-3.3mm}\includegraphics[height =0.40in]{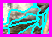} &
\hspace{-4.0mm}\includegraphics[height =0.40in]{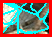} &
\hspace{-3.3mm}\includegraphics[height =0.40in]{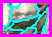} &
\hspace{-4.0mm}\includegraphics[height =0.40in]{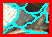} 
\end{tabular}
\begin{tabular}{cccccc}
%\hspace{-2mm}\includegraphics[height =0.41in]{figures/scheme1.pdf} &
% \hspace{-3.5mm}Input & GT segments & SLIC & SEAL & SSN & Ours\\
\hspace{-3.5mm}\includegraphics[height =0.84in]{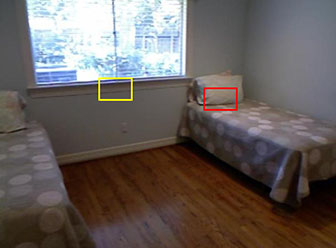}&
\hspace{-3.5mm}\includegraphics[height= 0.84in]{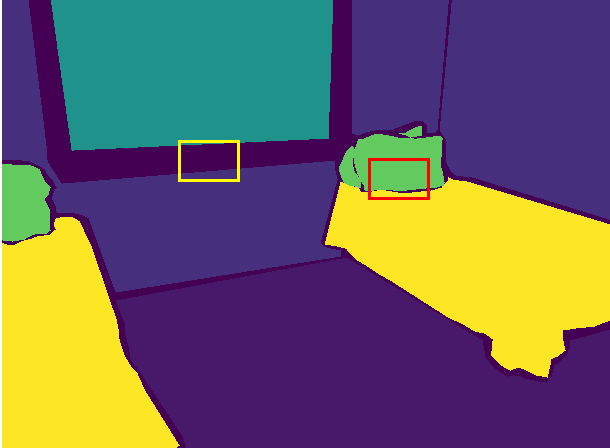} &
\hspace{-3.5mm}\includegraphics[height =0.84in]{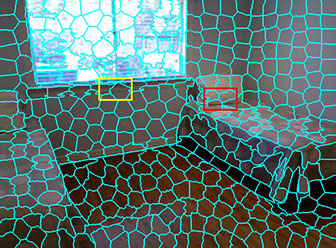} &
\hspace{-3.5mm}\includegraphics[height =0.84in]{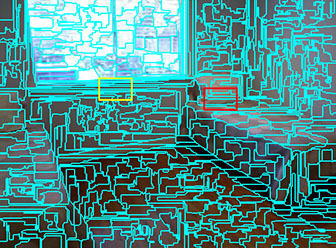} &
\hspace{-3.5mm}\includegraphics[height =0.84in]{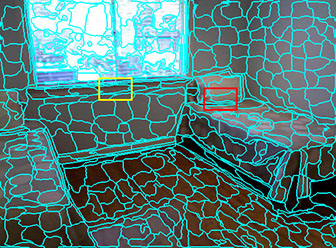} &
\hspace{-3.5mm}\includegraphics[height =0.84in]{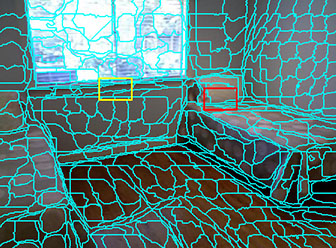} 
\end{tabular}
\begin{tabular}{cccccccccccc}
%\hspace{-2mm}\includegraphics[height =0.41in]{figures/scheme1.pdf} &
\hspace{-3.3mm}\includegraphics[height =0.38in]{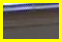} &
\hspace{-3.9mm}\includegraphics[height =0.38in]{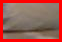} &
\hspace{-3.3mm}\includegraphics[height= 0.38in]{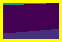} &
\hspace{-3.9mm}\includegraphics[height= 0.38in]{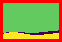} &
\hspace{-3.3mm}\includegraphics[height =0.38in]{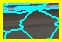} &
\hspace{-3.9mm}\includegraphics[height =0.38in]{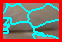} &
\hspace{-3.3mm}\includegraphics[height =0.38in]{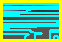} &
\hspace{-3.9mm}\includegraphics[height =0.38in]{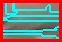} &
\hspace{-3.3mm}\includegraphics[height =0.38in]{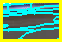} &
\hspace{-3.9mm}\includegraphics[height =0.38in]{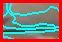} &
\hspace{-3.3mm}\includegraphics[height =0.38in]{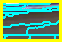} &
\hspace{-3.9mm}\includegraphics[height =0.38in]{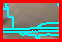} 
\end{tabular}
\caption{Example superpixel segmentation results. Compared to SEAL and SSN, our method is competitive or better in terms of object boundary adherence while generating more compact superpixels. {\bf Top rows}: BSDS500. {\bf Bottom rows}: NYUv2. }\vspace{-3mm}
\label{fig:spixel_viz}
\end{figure*}

\smallskip
\noindent{\bf Evaluation metrics.} We evaluate the superpixel methods using the popular metrics including achievable segmentation accuracy (ASA), boundary recall and precision (BR-BP), and compactness (CO). ASA quantifies the achievable accuracy for segmentation using the superpixels as pre-processing step, BR and BP measure the boundary adherence of superpixels given the ground truth, whereas CO assesses the compactness of superpixels. The higher these scores are, the better the segmentation result is. As in~\cite{StutzHL18}, for BR and BP evaluation, we set the boundary tolerance as 0.0025 times the image diagonal rounded to the closest integer. We refer readers to~\cite{StutzHL18} for the precise definitions.

\smallskip
\noindent{\bf Results on BSDS500.} BSDS500 contains 200 training, 100 validation, and 200 test images. As multiple labels are available for each image, we follow~\cite{JampaniSLYK18, Tu0JSC0K18} and treat each annotation as an individual sample, which results in 1633 training/validation samples and 1063 testing samples. 
%We first train our model on the training samples, and then tuning the hyper-parameters according to the validation performance, and finally train the model on the all then training and validation set with determined parameters. 
We train our model using both the training and validation samples.

Figure~\ref{fig:BSDS_res} reports the performance of all methods on BSDS500 test set. Our method outperforms all traditional methods on all evaluation metrics, except SLIC in term of CO. Comparing to the other deep learning-based methods, SEAL and SSN, our method achieves competitive or better results in terms of ASA and BR-BP, and significantly higher scores in term of CO. Figure~\ref{fig:spixel_viz} further shows example results of different methods. Note that, as discussed in~\cite{StutzHL18}, there is a well-known trade-off between boundary adherence and compactness. Although our method does not outperform existing methods on all the metrics, it appears to strike a better balance among them.  It is also worth noting that by achieving higher CO score, our method is able to better capture spatially coherent information and avoids paying too much attention to image details and noises. This characteristic tends to lead to better generalizability, as shown in the NYUv2 experiment results. 

% One thing worth to note here is compactness is not only a metrics for aesthetical consideration, but also for measuring the ability of the superpixel to capture spatially coherent information. Non-compact superpixels can be analogous to the over-fitting in machine learning, which use unnecessarily complicated decision boundary to separate data. This point can be verified in . When we directly apply the pre-trained model to NYUv2 dataset, the compactness of SSN start to approach our method, while the boundary precision-recall gap between the two methods disappears.

We also compare the runtime difference among deep learning-based (DL) methods. Figure~\ref{fig:time} reports the average runtime w.r.t. the number of generated superpixels on a NVIDIA GTX 1080Ti GPU device. Our method runs about 3 to 8 times faster than SSN and more than 50 times faster than SEAL. This is expected as our method uses a simple encoder-decoder network to directly generate superpixels, whereas SEAL and SSN first use deep networks to predict pixel affinity or features, and then apply traditional clustering methods (i.e., graph cuts or K-means) to get superpixels.

%==================== sPixel ========
% \begin{table}[htbp]
% \centering
% \caption{Average runtime (in second) of different DL methods.}
% \label{tab:sPixelTime}
% %\begin{small}
% \begin{tabular}{lcccc}
% \hline
% \#superpixels  & SEAL~\cite{Tu0JSC0K18}  & SSN~\cite{JampaniSLYK18}& Ours \\
% \hline 
% 200   & 1.130     &   0.089     &  0.011        \\
% 500   &  1.284       &     0.087       & 0.015       \\
% 1000  & 1.382 & 0.083  & 0.024\\
% \hline
% \end{tabular} \vspace{-3mm}
% %\end{small}
% \end{table}

%===========
\smallskip
\noindent{\bf Results on NYUv2.}  NYUv2 is a RGB-D dataset originally proposed for indoor scene understanding tasks, which contains 1,449 images with object instance labels. By removing the unlabelled regions near the image boundary,  \cite{StutzHL18} has developed a benchmark on a subset of 400 test images with size $608 \times 448$ for  superpixel evaluation. To test the generalizability of the learning-based methods, we directly apply the models of SEAL, SSN, and our method trained on BSDS500 to this dataset without any fine-tuning. 

Figure~\ref{fig:NYU_res} shows the performance of all methods on NYUv2. Generally, all deep learning-based methods perform well as they continue to achieve competitive or better performance against the traditional methods. Further, our method is shown to generalize better than SEAL and SSN, which is evident by comparing the corresponding curves in Figure~\ref{fig:BSDS_res} and~\ref{fig:NYU_res}. Specifically, our method outperforms SEAL and SSN in terms of BR-BP and CO, and is one of the best in terms of ASA.
% Specifically, our method outperforms SEAL and SSN in terms of both BR-BP and CO, and is second in terms of ASA, on the NYUv2 dataset. 
The visual results are shown in Figure~\ref{fig:spixel_viz}.   

\section{Application to Stereo Matching} 
\label{sec:disparity}
Stereo matching is a classic computer vision task which aims to find pixel correspondences between a pair of rectified images. Recent literature has shown that deep networks can boost the matching accuracy by building 4D cost volume (height$\times$width$\times$disparity$\times$feature channels) and aggregate the information using 3D convolution~\cite{chang2018PSMNet, abs-1810-02695, zhang2019ga}. However, such a design consumes large amounts of memory because of the extra ``disparity" dimension, limiting their ability to generate high-res outputs. A common remedy is to bilinearly upsample the predicted low-res disparity volumes for final disparity regression. As a result, object boundaries often become blur and fine details get lost.

In this section, we propose a downsampling/upsampling scheme based on the predicted superpixels and show how to integrate it into existing stereo matching pipelines to generate high-res outputs that better preserve the object boundaries and fine details.

\subsection{Network Design and Loss Function}
Figure~\ref{fig:framework} provides an overview of our method design. We choose PSMNet~\cite{chang2018PSMNet} as our task network. In order to incorporate our new downsampling/upsampling scheme, we change all the stride-2 convolutions in its feature extractor to stride-1, and remove the bilinear upsampling operations in the spatial dimensions. Given a pair of input images, we use our superpixel network to predict association maps $Q_l$, $Q_r$ and compute the superpixel center maps using Eq.~\eqref{eqn:center}. The center maps (\ie, downsampled images) are then fed into the modified PSMNet to get the low-res disparity volume. Next, the low-res volume is upsampled to original resolution with $Q_l$ according to Eq.~\eqref{eqn:pixel}, and the final disparity is computed using disparity regression. We refer readers to the supplementary materials for detailed specification. 

Same as PSMNet~\cite{chang2018PSMNet}, we use the 3-stage smooth $L_1$ loss with the weights $\alpha_1=0.5$, $\alpha_2=0.7$, and $\alpha_3=1.0$ for disparity prediction. And we use the SLIC loss (Eq.~\eqref{eqn:loss-slic}) for superpixel segmentation. The final loss function is:

 \begin{equation}
 \setlength{\abovedisplayskip}{-3pt}
L = \sum_{s=1} ^{3} \alpha_s \Big(  \frac{1}{N} \sum_{p=1}^{N} smooth_{L_1}(d_p - \hat{d}_p) \Big) + \frac{ \lambda}{N}  L_{SLIC}(Q)
\label{eqn:final}
\end{equation}
where $N$ is the total number of pixels, and $\lambda$ is a weight to balance the two terms. We set $\lambda=0.1$ for all experiments. 
\nop{
 \begin{equation}
    smooth_{L_1}(x) = 
    \begin{cases}
      0.5x^2 & \text{if $|x|<1$,}\\
      |x| - 0.5  & \text{otherwise}
    \end{cases}  
 \end{equation}
 }

% \begin{figure}[t]
% \centering
% \includegraphics[height =1.2in]{figures/stereo-net.pdf} 
% \caption{Overview of our network}
% \label{fig:arch}
% \vspace{-3mm}
% \end{figure}

\subsection{Experiments}
We have conducted experiments on three public datasets, SceneFlow~\cite{MayerIHFCDB16}, HR-VS~\cite{YangMHR19}, and Middlebury-v3~\cite{scharstein2014middlebury} to compared our model with PSMNet. To further verify the benefit of joint learning for superpixels and disparity estimation, we trained two different models for our method. In the first model {\bf Ours\_fixed}, we fix the parameters in superpixel network and train the rest of the network (\ie, the modified PSMNet) for disparity estimation. In the second model {\bf Ours\_joint}, we jointly train all networks in Figure~\ref{fig:framework}. 
For both models, the superpixel network is pre-trained on SceneFlow using the SLIC loss. The experiments are conducted on 4 Nvidia TITAN Xp GPUs. 

\begin{figure*}[t]
\centering
\begin{tabular}{ccccc}
%\hspace{-2mm}\includegraphics[height =0.41in]{figures/scheme1.pdf} &
\hspace{-2.7mm}Left image & \hspace{-2.5mm}GT disparity & \hspace{-2.5mm}PSMNet & \hspace{-2.5mm}Ours\_fixed & \hspace{-2.5mm} Ours\_joint\\
\hspace{-2.5mm}\includegraphics[height =0.74in]{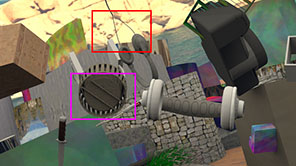} &
\hspace{-2.5mm}\includegraphics[height= 0.74in]{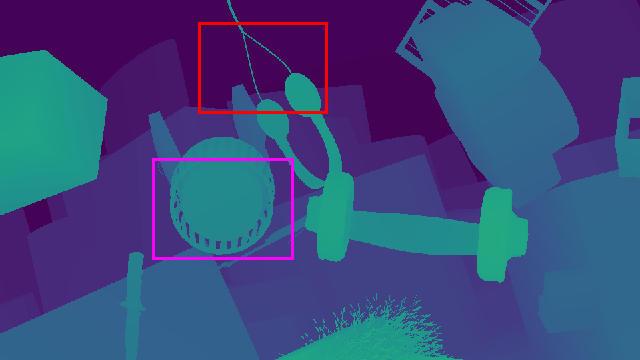} &
\hspace{-2.5mm}\includegraphics[height =0.74in]{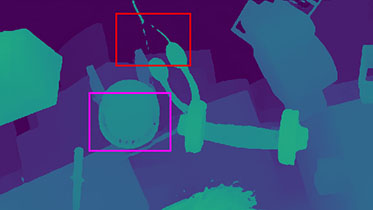} &
\hspace{-2.5mm}\includegraphics[height =0.74in]{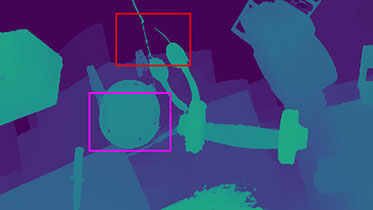} &
\hspace{-2.5mm}\includegraphics[height =0.74in]{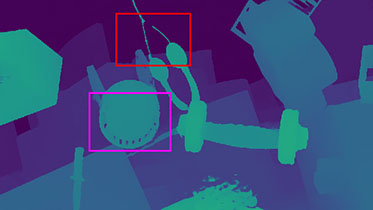} 
% \hspace{-3.5mm}\includegraphics[height =0.76in]{ICCV19-superpixel/figures/BSD_spixels/ours_268048.png} 
\end{tabular}
\begin{tabular}{cccccccccc}
%\hspace{-2mm}\includegraphics[height =0.41in]{figures/scheme1.pdf} &
\hspace{-2.6mm}\includegraphics[height =0.45in]{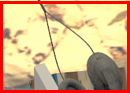} &
\hspace{-2.8mm}\includegraphics[height =0.45in]{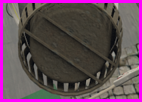} &
\hspace{-2.5mm}\includegraphics[height= 0.45in]{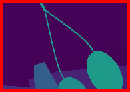}&
\hspace{-2.7mm}\includegraphics[height= 0.45in]{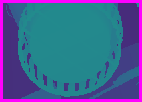} &
\hspace{-2.5mm}\includegraphics[height =0.45in]{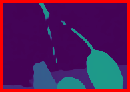} &
\hspace{-2.7mm}\includegraphics[height =0.45in]{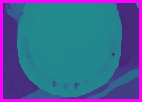} &
\hspace{-2.5mm}\includegraphics[height =0.45in]{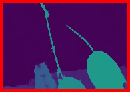} &
\hspace{-2.7mm}\includegraphics[height =0.45in]{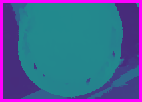} &
\hspace{-2.5mm}\includegraphics[height =0.45in]{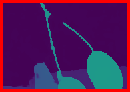} &
\hspace{-3.1mm}\includegraphics[height =0.45in]{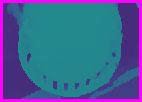}
\end{tabular}

\begin{tabular}{ccccc}
%\hspace{-2mm}\includegraphics[height =0.41in]{figures/scheme1.pdf} &
\hspace{-2.7mm}\includegraphics[height =0.76in]{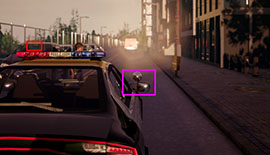} &
\hspace{-2.5mm}\includegraphics[height= 0.76in]{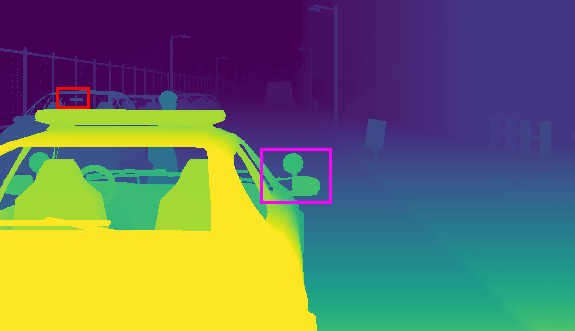} &
\hspace{-2.5mm}\includegraphics[height =0.76in]{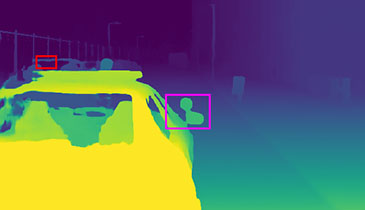} &
\hspace{-2.5mm}\includegraphics[height =0.76in]{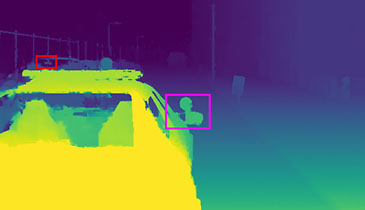} &
\hspace{-2.5mm}\includegraphics[height =0.76in]{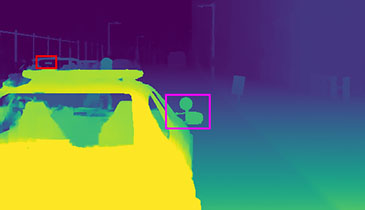} 
% \hspace{-3.5mm}\includegraphics[height =0.76in]{ICCV19-superpixel/figures/BSD_spixels/ours_268048.png} 
\end{tabular}

\begin{tabular}{cccccccccc}
%\hspace{-2mm}\includegraphics[height =0.41in]{figures/scheme1.pdf} &
\hspace{-2.6mm}\includegraphics[height =0.45in]{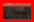} &
\hspace{-3.2mm}\includegraphics[height =0.45in]{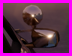} &
\hspace{-2.7mm}\includegraphics[height= 0.45in]{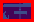}&
\hspace{-2.8mm}\includegraphics[height= 0.45in]{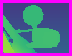} &
\hspace{-2.7mm}\includegraphics[height =0.45in]{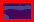} &
\hspace{-2.8mm}\includegraphics[height =0.45in]{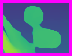} &
\hspace{-2.7 mm}\includegraphics[height =0.45in]{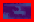} &
\hspace{-2.8mm}\includegraphics[height =0.45in]{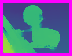} &
\hspace{-2.7mm}\includegraphics[height =0.45in]{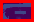} &
\hspace{-2.8mm}\includegraphics[height =0.45in]{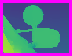}
\end{tabular}
\caption{Qualitative results on SceneFlow and HR-VS. Our method is able to better preserve fine details, such as the wires and mirror frameworks in the highlighted regions. \textbf{Top rows}: SceneFlow. \textbf{Bottom rows}: HR-VS.}\vspace{-2mm}
\label{fig:disp_viz}
\end{figure*}

\smallskip
\noindent{\bf Results on SceneFlow.}
SceneFlow is a synthetic dataset contains 35,454 training and 4,370 test frames with dense ground truth disparity. Following~\cite{chang2018PSMNet}, we exclude pixels with disparities greater than $192$ in training and test time. 

During training, we set $m=30$ in the SLIC loss and randomly crop the input images into size $ 512 \times 256 $.  To conduct 3D convolution at $1/4$ of the input resolution as PSMNet does, we predict superpixels with grid cell size $4\times4$ to perform the $4\times$ downsampling/upsampling. We train the model for 13 epochs with batch size 8. The initial learning rate is $1\times10^{-3}$, and is reduced to $5\times10^{-4}$ and $1\times10^{-4}$ after 11 and 12 epochs, respectively. For PSMNet, we use the authors' implementation and train it with the same learning schedule as our methods. % set to 

% We also conduct experiments to jointly train the two networks from scratch, and the performance is similar to ``Ours\_joint" setting. 

We use the standard end-point-error (EPE) as the evaluation metric, which measures the mean pixel-wise Euclidean distance between the predicted disparity and the ground truth. As shown in Table~\ref{tab:disp},  {\bf Ours\_joint} achieves the lowest EPE. Also note that {\bf Ours\_fixed} performs worse than the original PSMNet, which demonstrates the importance of joint training. Qualitative results are shown in Figure~\ref{fig:disp_viz}. One can see that both {\bf Ours\_fixed} and {\bf Ours\_joint} preserve fine details better than the original PSMNet.  

\smallskip
\noindent{\bf Results on HR-VS.}
HR-VS is a synthetic dataset with urban driving views. It contains 780 images at $2056\times 2464$ resolution. The valid disparity range is [9.66, 768]. Because no test set is released, we randomly choose 680 frames for training, and use the rest for testing. Due to the relatively small data size, we fine-tune all three models trained on SceneFlow in the previous experiment on this dataset. 
% Pixels with zero disparities are excluded in both training and test time.

Because of the high resolution and large disparity, the original PSMNet cannot be directly applied to the full size images. We follow the common practice to downsample both the input images and disparity maps to 1/4 size for training, and upsample the result to full resolution for evaluation. For our method,  we predict superpixels with grid cell size $16\times 16$ to perform $16\times$ downsampling/upsampling. During training, we set $ m=30 $, and randomly crop the images into size $ 2048 \times 1024 $. We train all methods for 200 epochs with batch size 4. The initial learning rate is  $1\times10^{-3}$ and reduced to $1\times10^{-4}$ after 150 epochs. 

%====================== disp ==========
\begin{table}[t]
\centering
\caption{End-point-error (EPE) on SceneFlow and HR-VS.}
\label{tab:disp}
\begin{tabular}{lcccccc}
\hline
Dataset & PSMNet~\cite{chang2018PSMNet} & Ours\_fixed &  Ours\_joint \\ %PSMNet~\cite{Chang18} &
\hline 
SceneFlow   & 1.04       &    1.07     & 0.93  \\ % &  1.09*
HR-VS   &   3.83   & 3.70 & 2.77 \\
\hline
\end{tabular}
\vspace{-3mm}
\end{table}

As shown in Table~\ref{tab:disp}, our models outperform the original PSMNet. And a significantly lower EPE is achieved by joint training. Note that, comparing to SceneFlow, we observe a larger performance gain on this high-res dataset, as we perform $16\times$ upsampling on HR-VS but only $4\times$ upsampling on SceneFlow. Qualitative results are shown in Figure~\ref{fig:disp_viz}.

\smallskip
\noindent{\bf Results on Middlebury-v3.}
Middlebury-v3 is a high-res real-world dataset with 10 training frames, 13 validation frames\footnote{Named as additional dataset in the official website.}, and 15 test frames. 
%As we cannot submit multiple results to the server, we only evaluate our joint training method. 
We use both training and validation frames to tune the {\bf Our\_joint} model pre-trained on SceneFlow with $16\times16$ superpixels. We set $m=60$ and train the model for 30 epochs with batch size 4. The initial learning rate is $1\times10^{-3}$ and divided by 10 after 20 epochs. 

% Table~\ref{tab:disp_mb} reports the results of our model on the official Middlebury-v3 leaderboard. 
Note that, for the experiment, our goal is not to achieve the highest rank on the official Middlebury-v3 leaderboard.
%since most top-performing methods on the leaderboard leverage additional data for training. 
But instead, to verify the effectiveness of the proposed superpixel-based downsample/upsampling scheme. Based on the leaderboard, our model outperforms PSMNet across all metrics, some of which are presented in Table~\ref{tab:disp_mb}. The results again verify the benefit of the proposed superpixel-based downsample/upsampling scheme.

% \begin{table*}[htbp]
% \centering
% \caption{Results on Middlebury-v3 benchmark where all pixels are evaluated.}
% \label{tab:disp}
% \begin{tabular}{lccccccccc}
% \hline
% Method & avgerr & rms & bad-1.0 & bad-2.0 & bad-4.0 & A99 & A95 & A90 & A50\\ %PSMNet~\cite{Chang18} &
% \hline 
% PSMNet_ROB~\cite{chang2018PSMNet}  & 8.78   & 23.3    &    67.3     & 47.2 & 29.2 & 106 & 44.3 & 22.8 & 2.20  \\ % &  1.09*
% Ours\_joint    &   7.11 & 19.1 & 67.0 & 46.8 & 27.5 & 94.5 & 37.9 & 13.8 & 1.98 \\
% \hline
% \end{tabular}
% \end{table*}

\begin{table}[htbp]
\centering
\vspace{-1mm}
\caption{Results on Middlebury-v3 benchmark.}
\label{tab:disp_mb}
\begin{tabular}{lccccccccc}
\hline
Method & avgerr & rms &   bad-4.0 & A90  \\ %PSMNet~\cite{Chang18} &
\hline 
PSMNet\_ROB~\cite{chang2018PSMNet}  & 8.78   & 23.3   & 29.2  & 22.8 \\ % &  1.09*
Ours\_joint    &   7.11 & 19.1 & 27.5 &  13.8 \\
\hline
\end{tabular}
\vspace{-4mm}
\end{table}

\section{Conclusion}
This paper has presented a simple fully convolutional network for superpixel segmentation. Experiments on benchmark datasets show that the proposed model is computationally efficient, and can consistently achieve the state-of-the-art performance with good generalizability. Further, we have demonstrated that higher disparity estimation accuracy can be obtained by using superpixels to preserve object boundaries and fine details in a popular stereo matching network. In the future, we plan to apply the proposed superpixel-based downsampling/upsampling scheme to other dense prediction tasks, such as object segmentation and optical flow estimation, and explore different ways to use superpixels in these applications.

\smallskip
\noindent{\bf Acknowledgement.} This work is supported in part by NSF award \#1815491 and a gift from Adobe.

\newpage

{\small
\bibliographystyle{ieee_fullname}
\bibliography{cvpr20-superpixel}
}

% \newpage
\appendix
%%%%%%%%% TITLE
% \title{Supplementary Materials for \\ Superpixel Segmentation with Fully Convolutional Networks}
\section{Supplementary Materials }
%%%%%%%%% ABSTRACT
%=========== supplement materials ========  
In Section~\ref{SpixelNet} and Section~\ref{DisparityNet}, we provide the detailed architecture designs for the superpixel segmentation network and the stereo matching network, respectively. In Section~\ref{add-viz-res}, we report additional qualitative results for superpixel segmentation on BSDS500 and NYUv2, disparity estimation on Sceneflow, HR-VS, and Middlebury-v3, and superpixel segmentation on HR-VS. 

\subsection{Superpixel Segmentation Network} \label{SpixelNet}
Table~\ref{tab:spixel-net} shows the specific design of our superpixel segmentation network. We use a standard encoder-decoder design with skip connections to predict the superpixel association map $Q$. Batch normalization and leaky Relu with negative slope 0.1 are used for all the convolution layers, except for the association prediction layer (assoc) where softmax is applied. 

\begin{table}[h]
\centering
\caption{ Specification of our superpixel segmentation network architecture.
%Unless otherwise stated, all the convolution operation follow with batch normalization and use leaky Relu with slope 0.1 as activation function.
}
\label{tab:spixel-net}
\begin{threeparttable}[h]
\addtolength{\tabcolsep}{-2pt}  
\scriptsize
\begin{tabular}{|c|ccc|cc|p{1.4cm}<{\centering}|}
\hline
Name 		& Kernel 		& Str. & Ch I/O   & InpRes 			& OutRes 		& Input \\
\hline 
cnv0a	    & $3\times 3$   &  1   & 3/16    & $208\times 208$ &  $208\times 208$ & image\\
cnv0b	    & $3\times 3$   &  1   & 16/16   & $208\times 208$  &  $208\times 208$ & cnv0a \\
cnv1a 		& $3\times 3$   &  2   & 16/32  & $208\times 208$  &  $104\times 104$ & cnv0b\\
cnv1b 		& $3\times 3$   &  1   & 32/32  & $104\times 104$  &  $104\times 104$ & cnv1a\\
cnv2a 		& $3\times 3$   &  2   & 32/64  & $104\times 104$  &  $52\times 52$ & cnv1b\\
cnv2b 		& $3\times 3$   &  1   & 64/64   & $52\times 52$   &  $52\times 52$ & cnv2a \\
cnv3a 		& $3\times 3$   &  2   & 64/128 & $52\times 52$  &  $26\times 26$ 	& cnv2b \\
cnv3b 		& $3\times 3$   &  1   & 128/128 & $26\times 26$  &  $26\times 26$ & cnv3a \\
cnv4a 		& $3\times 3$   &  2   & 128/256 & $26\times 26$  &  $13\times 13$ 	& cnv3b \\
cnv4b 		& $3\times 3$   &  1   & 256/256 & $13\times 13$  &   $13\times 13$   & cnv4a\\
upcnv3 		& $4\times 4$   &  2   & 256/128 & $13\times 13$  &   $26\times 26$ 	& cnv4b \\
icnv3 		& $3\times 3$   &  1   & 256/128 & $26\times 26$  &  $26\times 26$ 	& upcnv3+cnv3b \\
upcnv2 		& $4\times 4$   &  2   & 128/64 & $26\times 26$  &  $52\times 52$ 	& icnv3 \\
icnv2 		& $3\times 3$   &  1   & 128/64 & $52\times 52$  &  $52\times 52$ 	& upcnv2+cnv2b \\
upcnv1 		& $4\times 4$   &  2   & 64/32  & $52\times 52$  &  $104\times 104$ 	& icnv2 \\
icnv1 		& $3\times 3$   &  1   & 64/32 &  $104\times 104$  &  $104\times 104$ 	& upcnv1+cnv1b\\
upcnv0 		& $4\times 4$   &  2   & 32/16  & $104\times 104$  &  $208\times 208$ 	& icnv1 \\
icnv0 		& $3\times 3$   &  1   & 32/16 &  $208\times 208$  & $208\times 208$ 	& upcnv0+cnv0b\\
assoc		& $3\times 3$   &  1   & 16/9 &   $208\times 208$  & $208\times 208$  	& icnv0 \\
\hline
\end{tabular}

%\begin{tablenotes}
%     \item[1] No batch normalization and use softmax as activation function
%\end{tablenotes}

\end{threeparttable}%
\end{table}

\begin{table}[h]
\centering
\caption{ Specification of our stereo matching network (SP-PSMNet) architecture. 
%Unless otherwise stated, all the convolution operation use leaky Relu with slope 0.1 as activation function.
}
\linespread{1.1}
\label{tab:disp-net}
\begin{threeparttable}[h]
\addtolength{\tabcolsep}{-5pt}  
\scriptsize

\begin{tabular}{|c|c|c|p{1.0cm}<{\centering}|p{2.4cm}<{\centering}|}
\hline
Name 		        & Kernel 	   & Str.    & Input 	& OutDim \\		
\hline
\multicolumn{5}{|c|}{Input} \\
\hline
Img\_1/2      &           &       &         &  $ H \times W \times 3$  \\
\hline 
\multicolumn{5}{|c|}{{\bf Superpixel segmentation and superpixel-based downsampling}} \\
\hline
{assoc\_1/2} & {see Table~\ref{tab:spixel-net}} &  &  {Img\_1/2}  & $ {H \times W \times 9} $ \\
{sImg\_1/2}       & {assoc\_1/2}     & {4}        & {Img\_1/2} & ${\frac{1}{4}H \times \frac{1}{4}W \times 3 }$  \\
\hline
\multicolumn{5}{|c|}{PSMNet feature extractor} \\
\hline
\textbf{cnv0\_1} 	     & $\mathbf{ 3\times 3, 32} $   &  \textbf{1}         & \textbf{sImg\_1/2} & $ \mathbf{ \frac{1}{4}H \times \frac{1}{4}W  \times 32} $  \\
cnv0\_2	     & $3\times 3, 32$   &  1         & cnv0\_1   &  $\frac{1}{4}H \times \frac{1}{4}W  \times 32 $ \\
cnv0\_3         & $3\times 3, 32$   &  1         & cnv0\_2   &  $ \frac{1}{4}H \times \frac{1}{4}W  \times 32  $ \\
cnv1\_x         & $ \begin{bmatrix}
                 3\times3,32  \\
                 3\times3, 32
                \end{bmatrix}\times3 $   &  1 &cnv0\_3  & $ \frac{1}{4}H \times \frac{1}{4}W  \times 32 $ \\
\textbf{ conv2\_x}        & $ \mathbf{ \begin{bmatrix}
                 \mathbf{ 3\times3, 64} \\
                 \mathbf{ 3\times3, 64}
                \end{bmatrix}\times16} $   &  \textbf{1} & \textbf{cnv1\_x} & $ \mathbf{\frac{1}{4}H \times \frac{1}{4}W  \times 64} $  \\

cnv3\_x         & $ \begin{bmatrix}
                 3\times3, 128 \\
                 3\times3, 128
                \end{bmatrix}\times3 $   &  1 &cnv2\_x  & $ \frac{1}{4}H \times \frac{1}{4}W  \times 128 $  \\
                
cnv4\_x         & $ \begin{bmatrix}
                 3\times3,128 \\
                 3\times3, 128
                \end{bmatrix}\times3, dila=2 $   &  1 &cnv3\_x  & $ \frac{1}{4}H \times \frac{1}{4}W  \times 128 $ \\
\hline

\multicolumn{5}{|c|}{PSMNet SPP module, cost volume, and 3D CNN} \\
\hline
output\_1 & 
\multicolumn{3}{c|}{\multirow{3}{*}{\tabincell{c}{Please refer to~\cite{chang2018PSMNet} for details}}} & $\frac{1}{4}H \times \frac{1}{4}W \times \frac{1}{4} D\times 1 $ \\
output\_2 & \multicolumn{3}{c|}{} & $\frac{1}{4}H \times \frac{1}{4}W \times \frac{1}{4} D\times 1 $ \\
output\_3 & \multicolumn{3}{c|}{} & $\frac{1}{4}H \times \frac{1}{4}W \times \frac{1}{4} D\times 1 $ \\
\hline
\multicolumn{5}{|c|}{{\bf Superpixel-based upsampling}} \\
\hline
\multirow{2}{*}{disp\_prb1} & bilinear upsampling & N.A.     & \multirow{2}{*}{output\_1 } & $\frac{1}{4}H \times \frac{1}{4}W \times D$ \\
  & assoc\_1              & 4     &  & $ H \times W \times D$\\
\hline
\multirow{2}{*}{disp\_prb2}& bilinear upsampling & N.A. &   \multirow{2}{*}{output\_2 } & $\frac{1}{4}H \times \frac{1}{4}W \times D$ \\
   & assoc\_1              & 4 &    & $ H \times W \times D$\\
\hline
\multirow{2}{*}{disp\_prb3}& bilinear upsampling & N.A. &  \multirow{2}{*}{output\_3} & $\frac{1}{4}H \times \frac{1}{4}W \times D$ \\
& assoc\_1             & 4 &   & $ H \times W \times D$\\
\hline
\multicolumn{5}{|c|}{PSMNet disparity regression} \\
\hline
disp\_1      & disparity regression & N.A. & disp\_prb1 & $ H \times W $ \\
disp\_2     & disparity regression & N.A. & disp\_prb2 & $ H \times W $ \\
disp\_3      & disparity regression & N.A. & disp\_prb3 & $ H \times W $\\
\hline
% \multicolumn{7}{|c|}{build the cost volume and the rest are same as PSMNet except SPPooling upsampling} \\
% \hline
\end{tabular}

\end{threeparttable}%
\end{table}

\subsection{Stereo Matching Network} \label{DisparityNet}

Table~\ref{tab:disp-net} shows the architecture design of stereo matching network, in which we modify PSMNet~\cite{chang2018PSMNet} to perform superpixel-based downsampling/upsampling operations. We name it superpixel-based PSMNet (SPPSMNet). The layers which are different from the orignal PSMNet have been highlighted in bold face. In Table~\ref{tab:disp-net}, we use input image size $256\times 512$ with maximum disparity  $D=192$, which is the same as the original PSMNet, and we set superpixel grid cell size $4\times4$ to perform $4\times$ downsampling/upsampling.

For stereo matching tasks with high resolution images (\ie, HR-VS and Middilebury-v3), we use input image size $1024\times 2048$ with maximum disparity $D=768$, and we set superpixel grid cell size $16\times 16$ to perform $16\times$ downsampling/upsampling. To further reduce the GPU memory usage, in the high-res stereo matching tasks, we reduce the channel number of the layers ``cnv4a" and ``cnv4b" in the superpixel segmentation network from $256$ to $128$, remove the batch normalization operation in the superpixel segmentation network, and perform superpixel-based spatial upsampling after the disparity regression.

% The superpixel segmentation network is exactly same as the one described in Sec.~\ref{SpixelNet}, except for layer ``conv4b" where we change the channel number from 256 to 128  to save the memory. For the PSMNet part, we change the stride-2 convolution in original feature extractor to stride 1 and replace the bilinear cost upsampling with our supeprixel based upsampling. The setting of the activation functions is the same as the original PSMNet. 

\subsection{Additional Qualitative Results} 
\label{add-viz-res}

\subsubsection{Superpixel Segmentation}
\begin{figure*}[t]
\centering
\begin{tabular}{cccccc}
%\hspace{-2mm}\includegraphics[height =0.41in]{figures/scheme1.pdf} &
Input &\hspace{-3.5mm}GT segments & \hspace{-3.5mm}SLIC & \hspace{-3.5mm}SEAL & \hspace{-3.5mm}SSN & \hspace{-3.5mm}Ours \\
\includegraphics[height =0.70in]{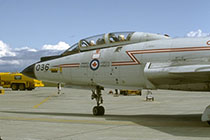} &
\hspace{-3.5mm}\includegraphics[height= 0.692in]{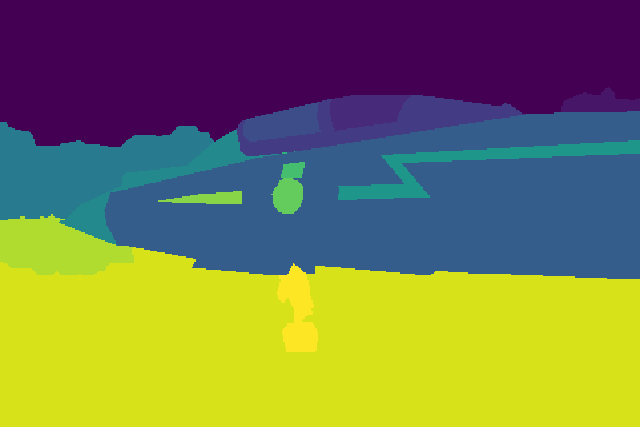} &
\hspace{-3.5mm}\includegraphics[height =0.692in]{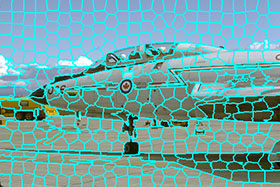} &
\hspace{-3.5mm}\includegraphics[height =0.692in]{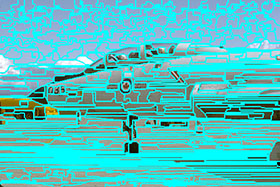} &
\hspace{-3.5mm}\includegraphics[height =0.692in]{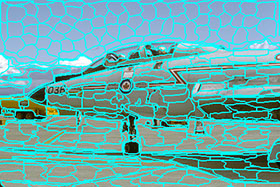} &
\hspace{-3.5mm}\includegraphics[height =0.692in]{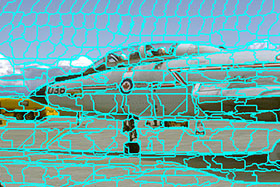} \\

\includegraphics[height =0.70in]{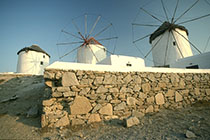} &
\hspace{-3.5mm}\includegraphics[height= 0.692in]{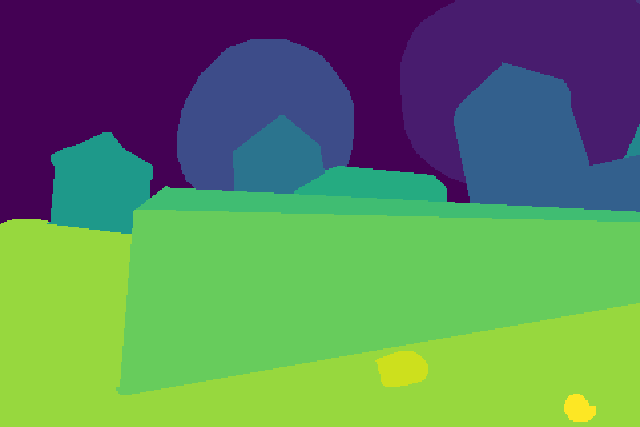} &
\hspace{-3.5mm}\includegraphics[height =0.692in]{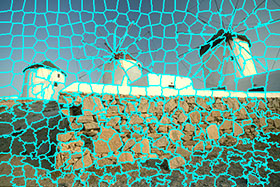} &
\hspace{-3.5mm}\includegraphics[height =0.692in]{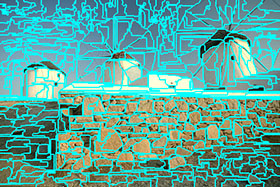} &
\hspace{-3.5mm}\includegraphics[height =0.692in]{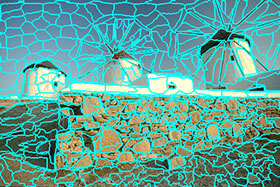} &
\hspace{-3.5mm}\includegraphics[height =0.692in]{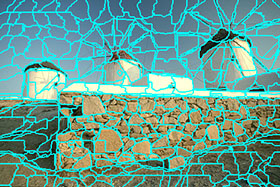} \\

\includegraphics[height =0.70in]{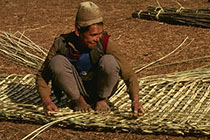} &
\hspace{-3.5mm}\includegraphics[height= 0.692in]{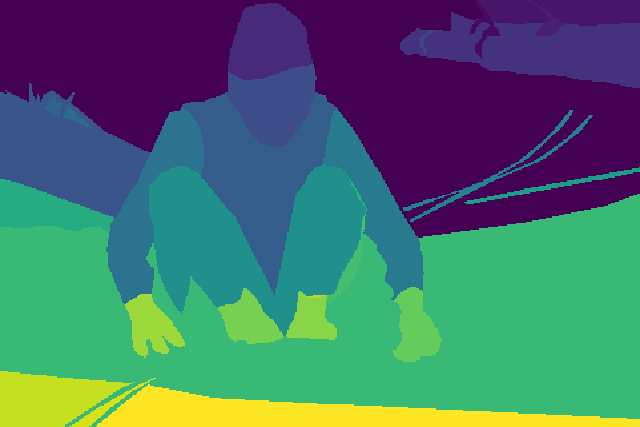} &
\hspace{-3.5mm}\includegraphics[height =0.692in]{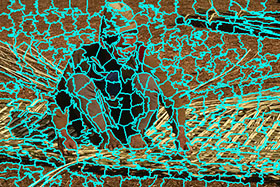} &
\hspace{-3.5mm}\includegraphics[height =0.692in]{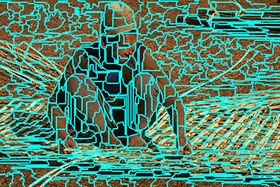} &
\hspace{-3.5mm}\includegraphics[height =0.692in]{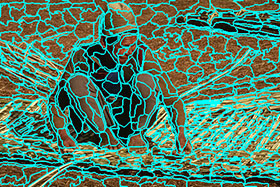} &
\hspace{-3.5mm}\includegraphics[height =0.692in]{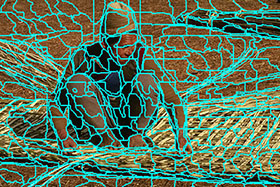} \\

\includegraphics[height =0.70in]{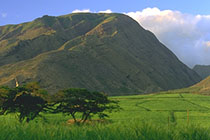} &
\hspace{-3.5mm}\includegraphics[height= 0.692in]{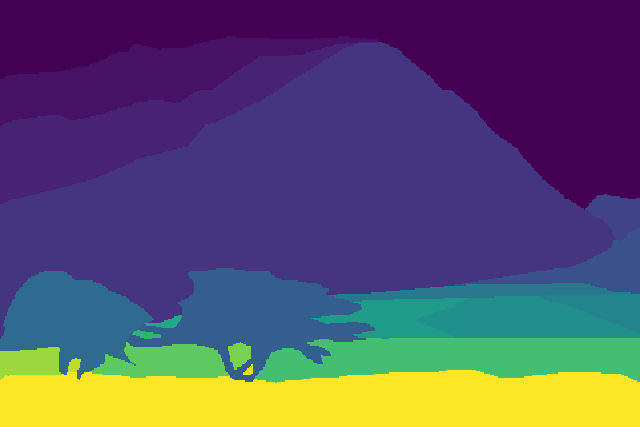} &
\hspace{-3.5mm}\includegraphics[height =0.692in]{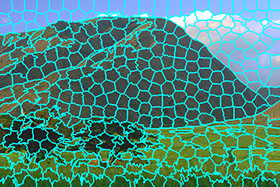} &
\hspace{-3.5mm}\includegraphics[height =0.692in]{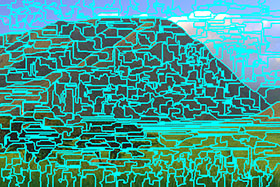} &
\hspace{-3.5mm}\includegraphics[height =0.692in]{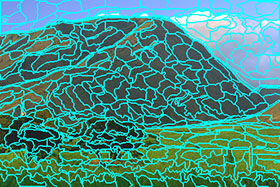} &
\hspace{-3.5mm}\includegraphics[height =0.692in]{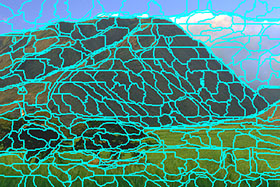} \\

\includegraphics[height =0.70in]{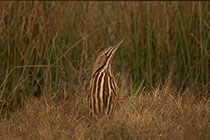} &
\hspace{-3.5mm}\includegraphics[height= 0.692in]{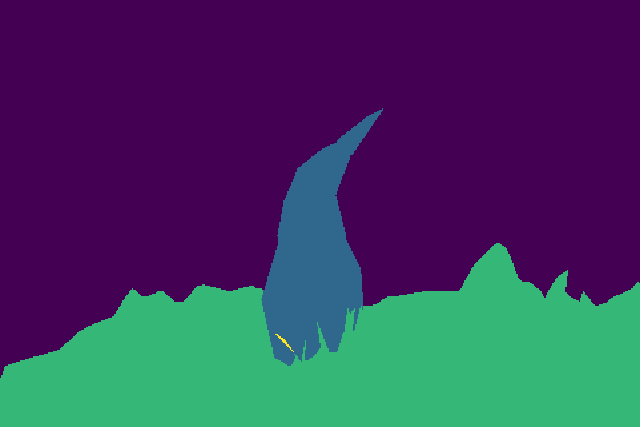} &
\hspace{-3.5mm}\includegraphics[height =0.692in]{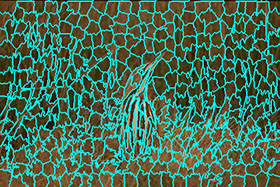} &
\hspace{-3.5mm}\includegraphics[height =0.692in]{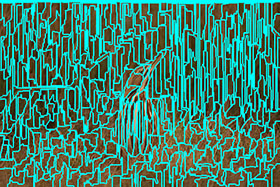} &
\hspace{-3.5mm}\includegraphics[height =0.692in]{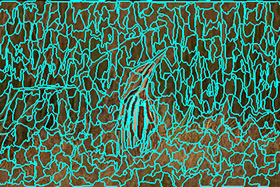} &
\hspace{-3.5mm}\includegraphics[height =0.692in]{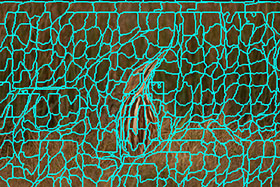} \\

\end{tabular}
\caption{Additional superpixel segmentation results on BSDS500. 
%Compared to SEAL and SSN, our method is competitive or better in terms of object boundary adherence while generating more compact superpixels. 
}
\label{fig:spixel_viz_supp_bsds}
\end{figure*}

\begin{figure*}[hb]
\centering
\begin{tabular}{cccccc}
%\hspace{-2mm}\includegraphics[height =0.41in]{figures/scheme1.pdf} &
Input &\hspace{-3.5mm}GT segments & \hspace{-3.5mm}SLIC & \hspace{-3.5mm}SEAL & \hspace{-3.5mm}SSN & \hspace{-3.5mm}Ours \\
\includegraphics[height =0.77in]{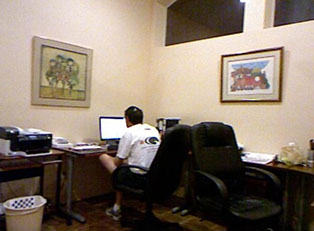} &
\hspace{-3.5mm}\includegraphics[height= 0.77in]{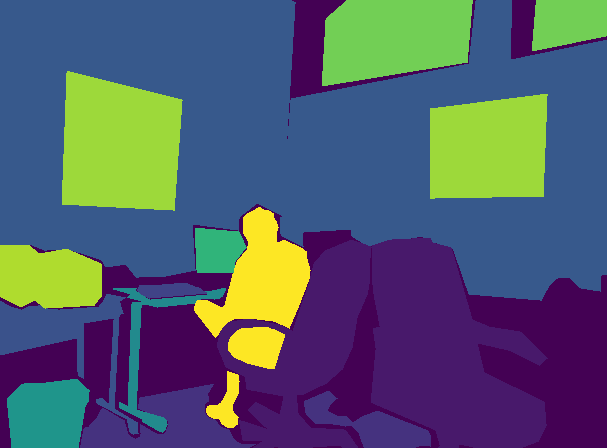} &
\hspace{-3.5mm}\includegraphics[height =0.77in]{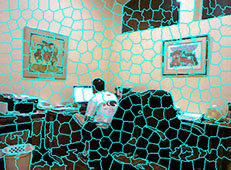} &
\hspace{-3.5mm}\includegraphics[height =0.77in]{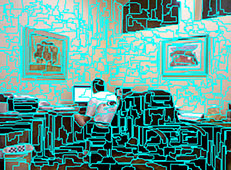} &
\hspace{-3.5mm}\includegraphics[height =0.77in]{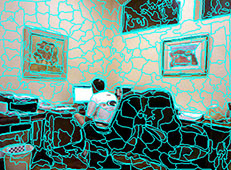} &
\hspace{-3.5mm}\includegraphics[height =0.77in]{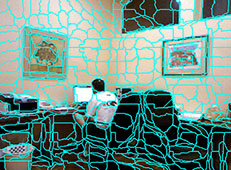} \\

\includegraphics[height =0.77in]{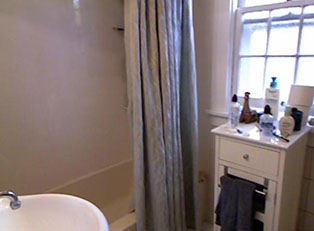} &
\hspace{-3.5mm}\includegraphics[height= 0.77in]{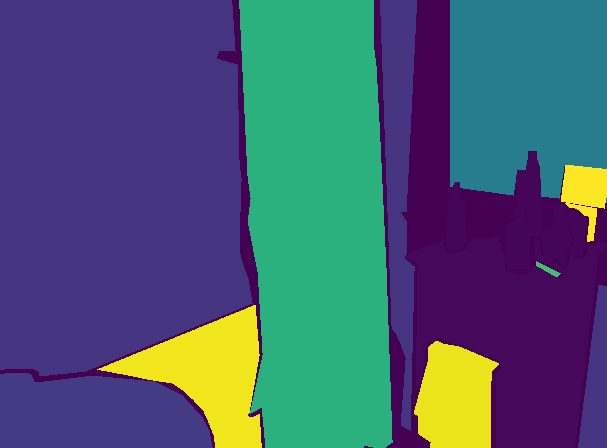} &
\hspace{-3.5mm}\includegraphics[height =0.77in]{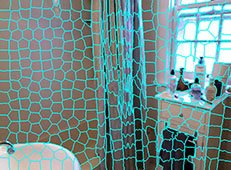} &
\hspace{-3.5mm}\includegraphics[height =0.77in]{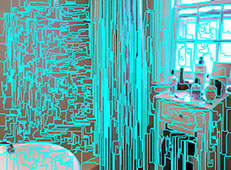} &
\hspace{-3.5mm}\includegraphics[height =0.77in]{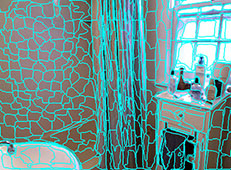} &
\hspace{-3.5mm}\includegraphics[height =0.77in]{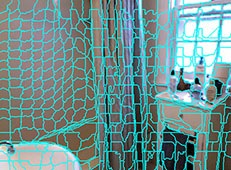} \\

\includegraphics[height =0.77in]{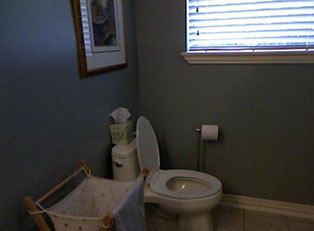} &
\hspace{-3.5mm}\includegraphics[height= 0.77in]{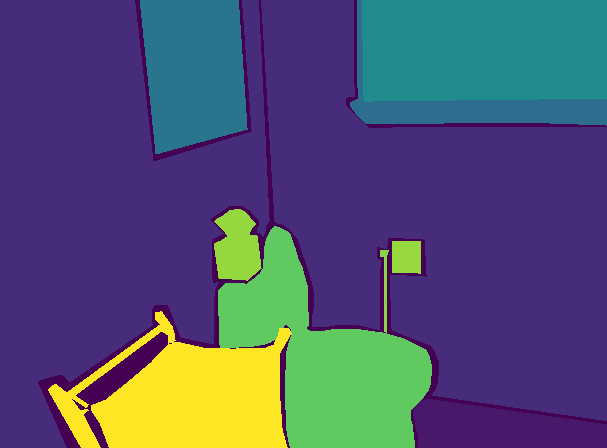} &
\hspace{-3.5mm}\includegraphics[height =0.77in]{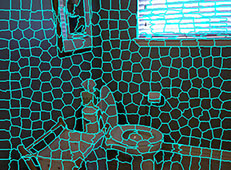} &
\hspace{-3.5mm}\includegraphics[height =0.77in]{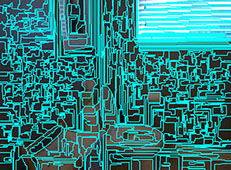} &
\hspace{-3.5mm}\includegraphics[height =0.77in]{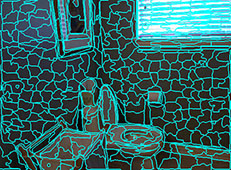} &
\hspace{-3.5mm}\includegraphics[height =0.77in]{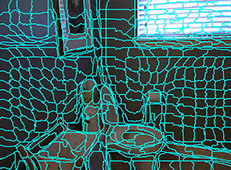} \\

\includegraphics[height =0.77in]{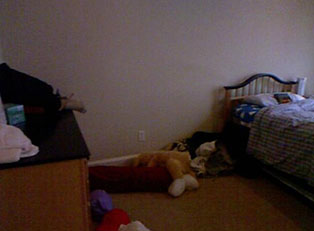} &
\hspace{-3.5mm}\includegraphics[height= 0.77in]{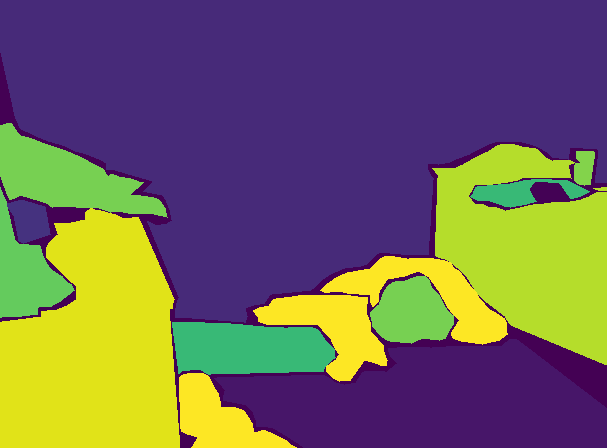} &
\hspace{-3.5mm}\includegraphics[height =0.77in]{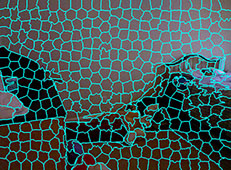} &
\hspace{-3.5mm}\includegraphics[height =0.77in]{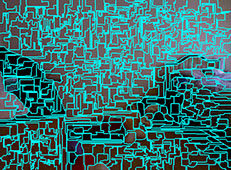} &
\hspace{-3.5mm}\includegraphics[height =0.77in]{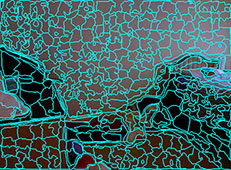} &
\hspace{-3.5mm}\includegraphics[height =0.77in]{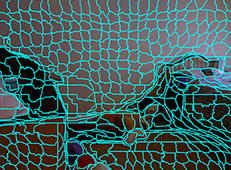} \\

\includegraphics[height =0.77in]{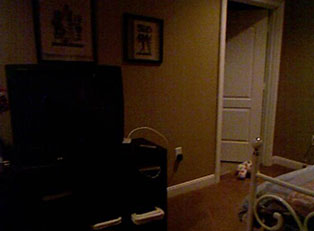} &
\hspace{-3.5mm}\includegraphics[height= 0.77in]{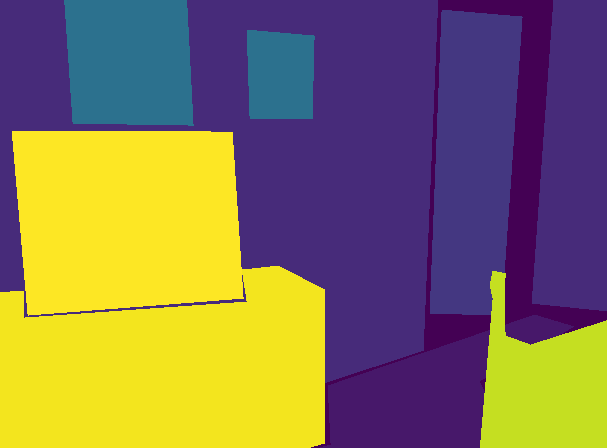} &
\hspace{-3.5mm}\includegraphics[height =0.77in]{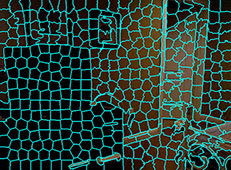} &
\hspace{-3.5mm}\includegraphics[height =0.77in]{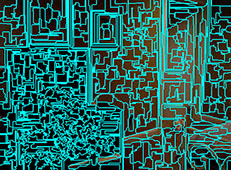} &
\hspace{-3.5mm}\includegraphics[height =0.77in]{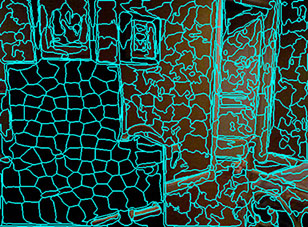} &
\hspace{-3.5mm}\includegraphics[height =0.77in]{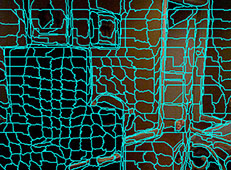} \\

\end{tabular}
\caption{Additional superpixel segmentation results on NYUv2.
%Compared to SEAL and SSN, our method is competitive or better in terms of object boundary adherence while generating more compact superpixels.
}
\label{fig:spixel_viz_supp_nyu}
\end{figure*}

Figure \ref{fig:spixel_viz_supp_bsds} and Figure \ref{fig:spixel_viz_supp_nyu} show additional qualitative results for superpixel segmentation on BSDS500 and NYUv2. The three learning-based methods, SEAL, SSN, and ours, can recover more detailed boundaries than SLIC, such as the hub of the windmill in the second row of Figure~\ref{fig:spixel_viz_supp_bsds} and the pillow on the right bed in the fourth row of Figure~\ref{fig:spixel_viz_supp_nyu}. Compared to SEAL and SSN, our method usually generate more compact superpixels.

\subsubsection{Application to Stereo Matching}
\begin{figure*}[t!]
\centering
\begin{tabular}{cccc}
\hspace{-3mm}Left image/GT & \hspace{-3mm}PSMNet & \hspace{-3mm}Ours\_fixed & \hspace{-3mm}Ours\_joint \\

% \hspace{-3mm}image/GT & SPS-Stereo & DispNetC & Ours \\
\hspace{-3mm}\includegraphics[height=0.9in]{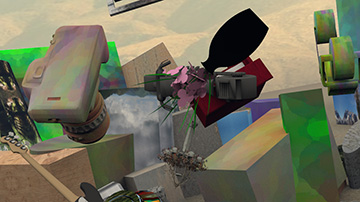}  &
\hspace{-3mm}\includegraphics[height=0.9in]{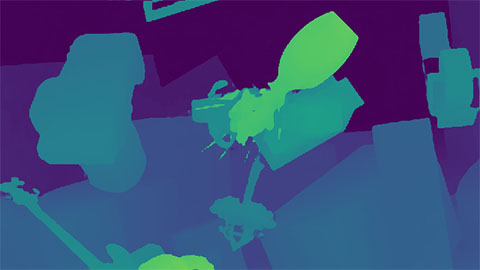}&
\hspace{-3mm}\includegraphics[height=0.9in]{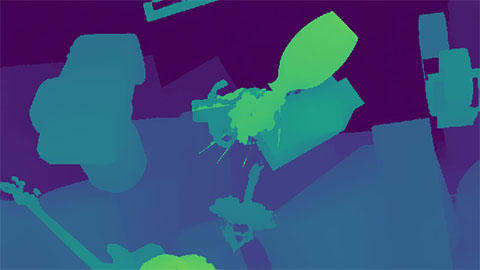} &
\hspace{-3mm}\includegraphics[height=0.9in]{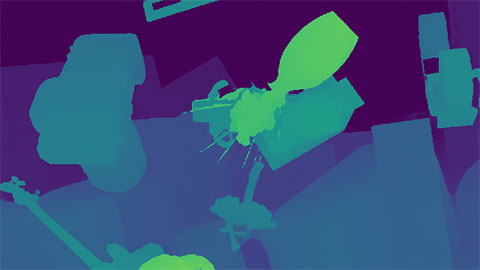} \\
\hspace{-3mm}\includegraphics[height=0.9in]{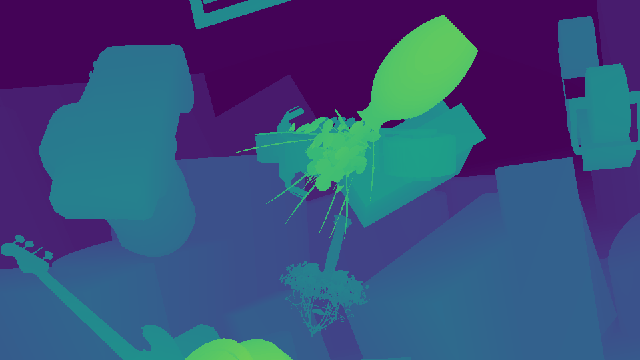} &
\hspace{-3mm}\includegraphics[height=0.9in]{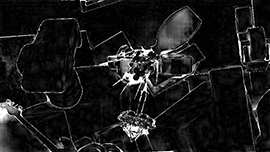} &
\hspace{-3mm}\includegraphics[height=0.9in]{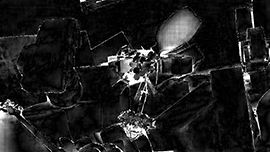} &
\hspace{-3mm}\includegraphics[height=0.9in]{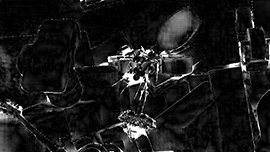} \\
% \arrayrulecolor{red} \hline \\ [-1.8ex]

\hspace{-3mm}\includegraphics[height=0.9in]{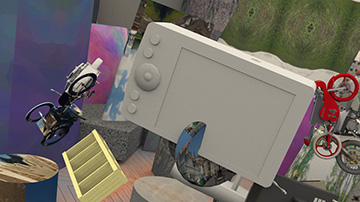}  &
\hspace{-3mm}\includegraphics[height=0.9in]{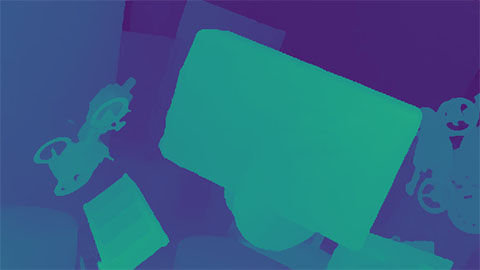}&
\hspace{-3mm}\includegraphics[height=0.9in]{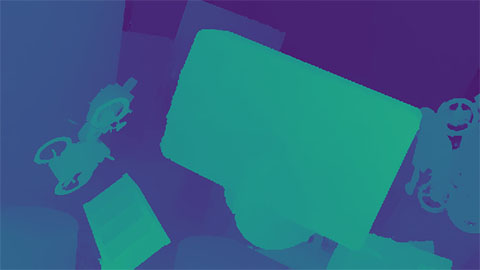} &
\hspace{-3mm}\includegraphics[height=0.9in]{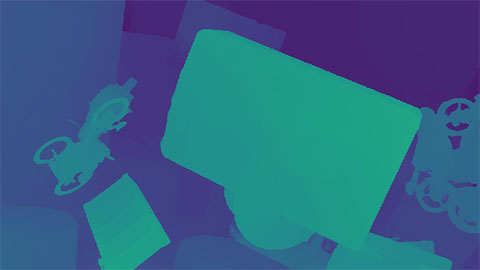} \\
\hspace{-3mm}\includegraphics[height=0.9in]{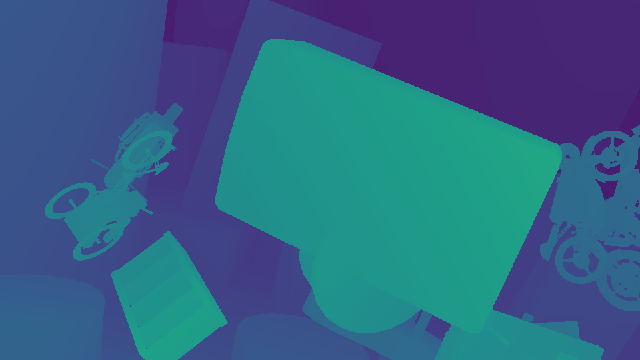} &
\hspace{-3mm}\includegraphics[height=0.9in]{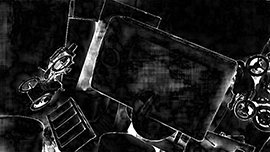} &
\hspace{-3mm}\includegraphics[height=0.9in]{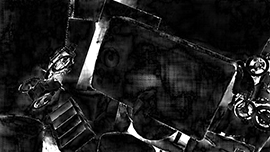} &
\hspace{-3mm}\includegraphics[height=0.9in]{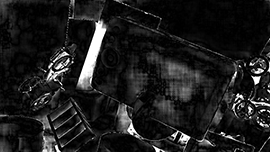} \\
% % \arrayrulecolor{red} \hline \\ [-1.8ex]

% \hspace{-3mm}image/GT & SPS-Stereo & DispNetC & Ours \\
\hspace{-3mm}\includegraphics[height=0.9in]{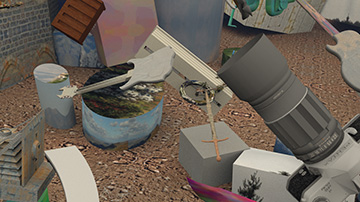}  &
\hspace{-3mm}\includegraphics[height=0.9in]{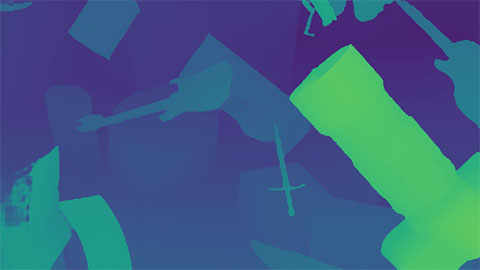}&
\hspace{-3mm}\includegraphics[height=0.9in]{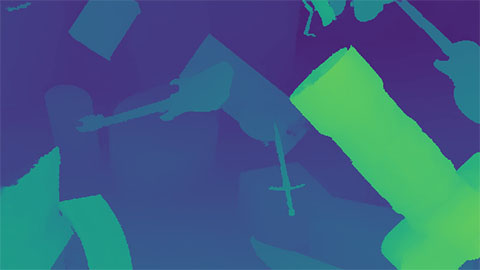} &
\hspace{-3mm}\includegraphics[height=0.9in]{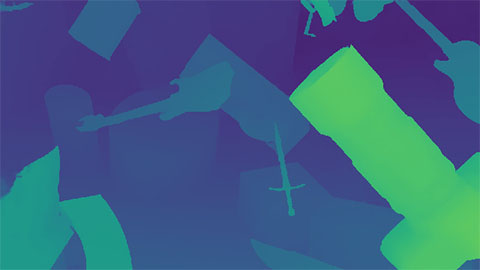} \\
\hspace{-3mm}\includegraphics[height=0.9in]{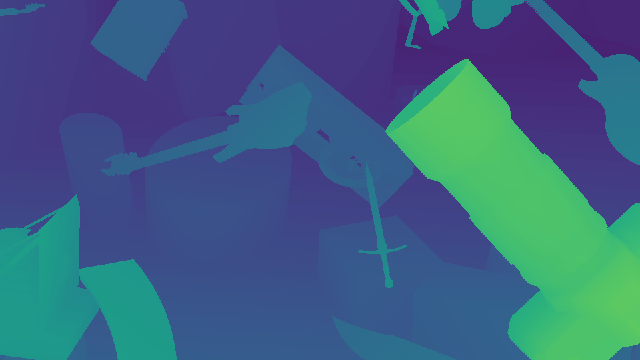} &
\hspace{-3mm}\includegraphics[height=0.9in]{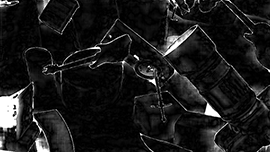} &
\hspace{-3mm}\includegraphics[height=0.9in]{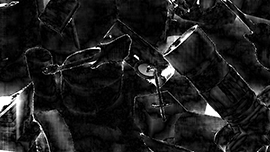} &
\hspace{-3mm}\includegraphics[height=0.9in]{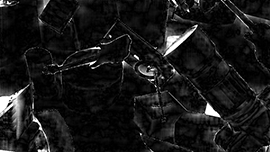} \\
% % \arrayrulecolor{red} \hline \\ [-1.8ex]

% % \hspace{-3mm}image/GT & SPS-Stereo & DispNetC & Ours \\
\hspace{-3mm}\includegraphics[height=0.9in]{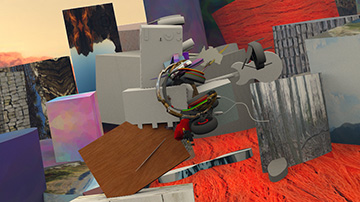}  &
\hspace{-3mm}\includegraphics[height=0.9in]{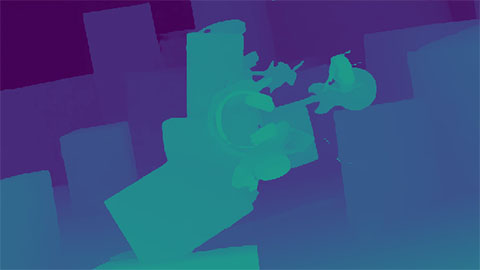}&
\hspace{-3mm}\includegraphics[height=0.9in]{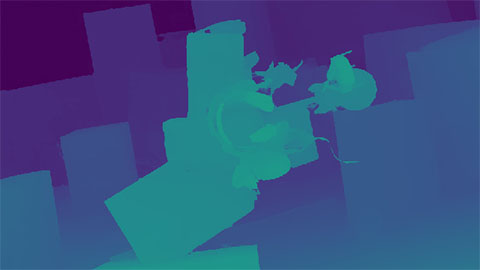} &
\hspace{-3mm}\includegraphics[height=0.9in]{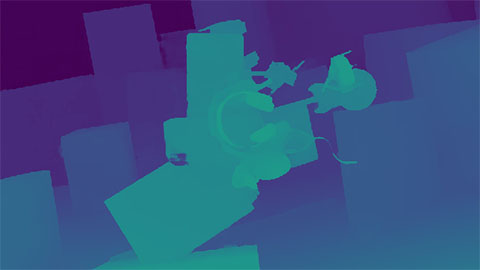} \\
\hspace{-3mm}\includegraphics[height=0.9in]{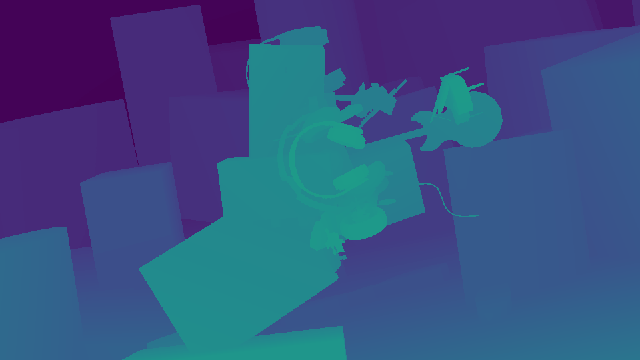} &
\hspace{-3mm}\includegraphics[height=0.9in]{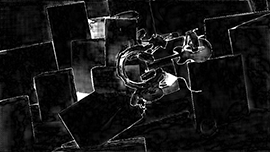} &
\hspace{-3mm}\includegraphics[height=0.9in]{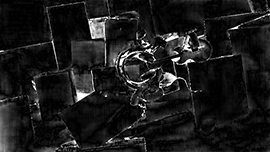} &
\hspace{-3mm}\includegraphics[height=0.9in]{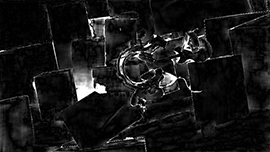} \\

\end{tabular}
\caption{Disparity prediction results on SceneFlow. For each method, we show both the predicted disparity map (top) and the error map (bottom). For the error map, the darker the color, the lower the end point error (EPE).
%For error map, the pixel value is set to 255 (white) if the EPE is larger than 10, and the darker the color, the lower the EPE. 
}
\label{fig:disp_sceneflow}
\end{figure*}

\begin{figure*}[t!]
\centering
\begin{tabular}{cccc}
\hspace{-3mm}Left image/GT & \hspace{-3mm}PSMNet & \hspace{-3mm}Ours\_fixed & \hspace{-3mm}Ours\_joint \\

% \hspace{-3mm}image/GT & SPS-Stereo & DispNetC & Ours \\
\hspace{-3mm}\includegraphics[height=0.95in]{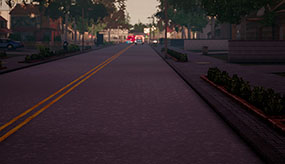}  &
\hspace{-3mm}\includegraphics[height=0.95in]{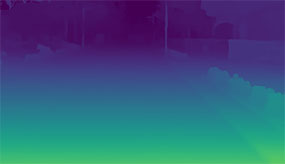}&
\hspace{-3mm}\includegraphics[height=0.95in]{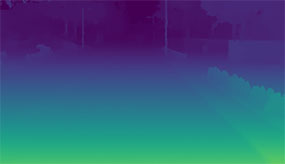} &
\hspace{-3mm}\includegraphics[height=0.95in]{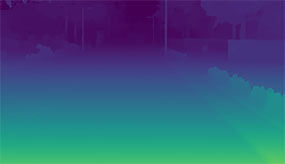} \\
\hspace{-3mm}\includegraphics[height=0.95in]{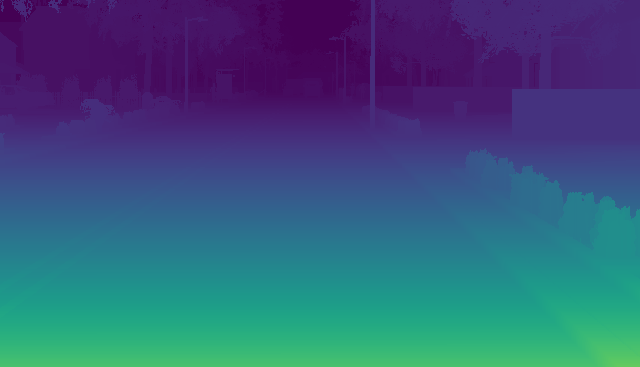} &
\hspace{-3mm}\includegraphics[height=0.95in]{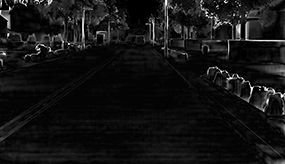} &
\hspace{-3mm}\includegraphics[height=0.95in]{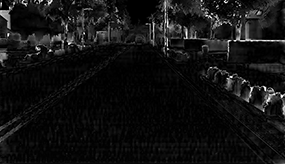} &
\hspace{-3mm}\includegraphics[height=0.95in]{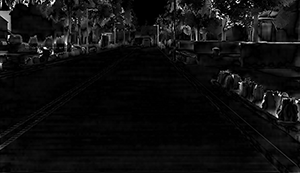} \\
% \arrayrulecolor{red} \hline \\ [-1.8ex]

\hspace{-3mm}\includegraphics[height=0.95in]{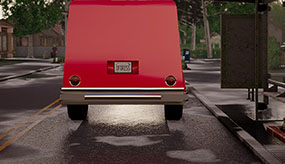}  &
\hspace{-3mm}\includegraphics[height=0.95in]{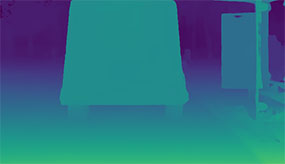}&
\hspace{-3mm}\includegraphics[height=0.95in]{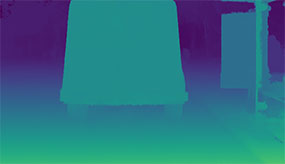} &
\hspace{-3mm}\includegraphics[height=0.95in]{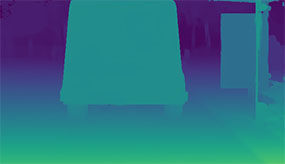} \\
\hspace{-3mm}\includegraphics[height=0.95in]{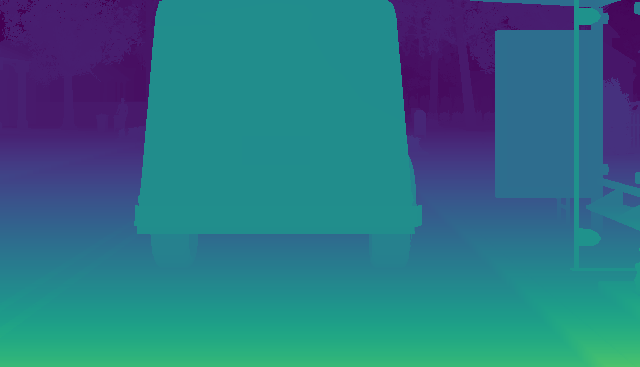}  &
\hspace{-3mm}\includegraphics[height=0.95in]{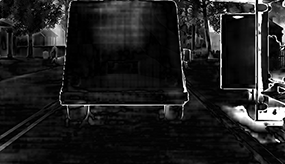} &
\hspace{-3mm}\includegraphics[height=0.95in]{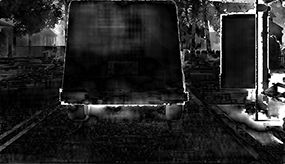} &
\hspace{-3mm}\includegraphics[height=0.95in]{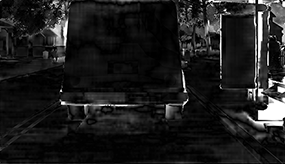} \\
% % % \arrayrulecolor{red} \hline \\ [-1.8ex]

% % \hspace{-3mm}image/GT & SPS-Stereo & DispNetC & Ours \\
\hspace{-3mm}\includegraphics[height=0.95in]{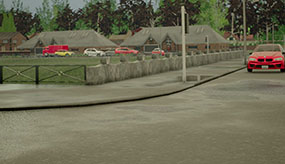}  &
\hspace{-3mm}\includegraphics[height=0.95in]{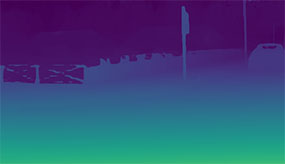}&
\hspace{-3mm}\includegraphics[height=0.95in]{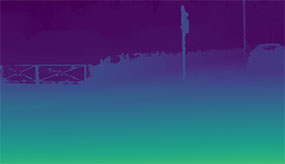} &
\hspace{-3mm}\includegraphics[height=0.95in]{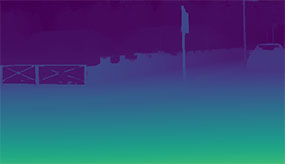} \\
\hspace{-3mm}\includegraphics[height=0.95in]{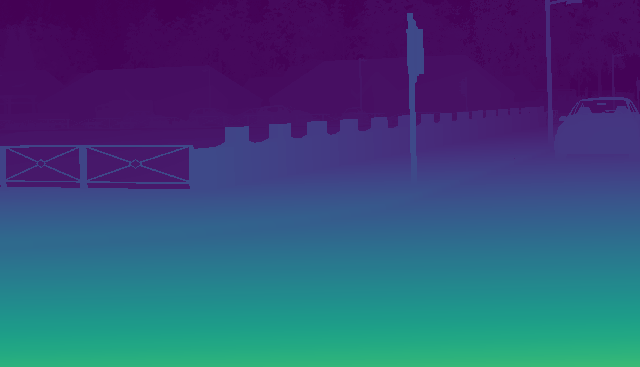} &
\hspace{-3mm}\includegraphics[height=0.95in]{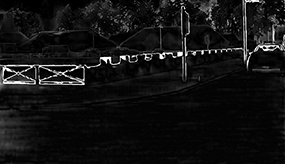} &
\hspace{-3mm}\includegraphics[height=0.95in]{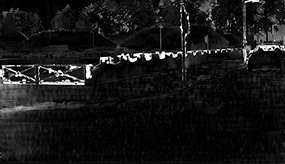} &
\hspace{-3mm}\includegraphics[height=0.95in]{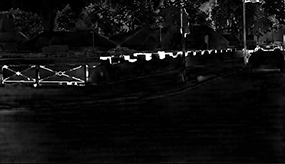}\\
% % % \arrayrulecolor{red} \hline \\ [-1.8ex]

% % % \hspace{-3mm}image/GT & SPS-Stereo & DispNetC & Ours \\
\hspace{-3mm}\includegraphics[height=0.95in]{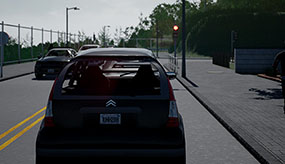}  &
\hspace{-3mm}\includegraphics[height=0.95in]{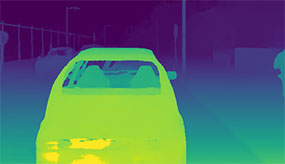}&
\hspace{-3mm}\includegraphics[height=0.95in]{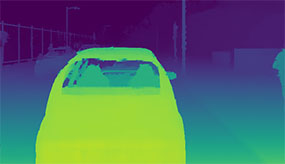} &
\hspace{-3mm}\includegraphics[height=0.95in]{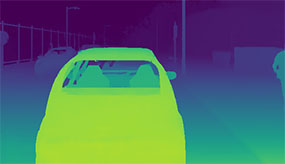} \\
\hspace{-3mm}\includegraphics[height=0.95in]{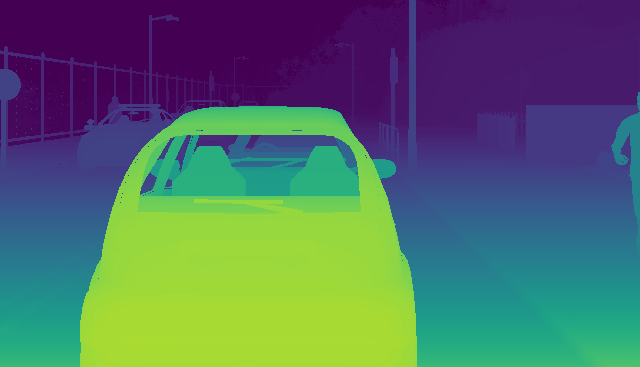} &
\hspace{-3mm}\includegraphics[height=0.95in]{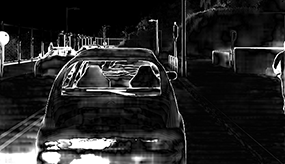} &
\hspace{-3mm}\includegraphics[height=0.95in]{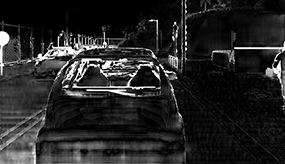} &
\hspace{-3mm}\includegraphics[height=0.95in]{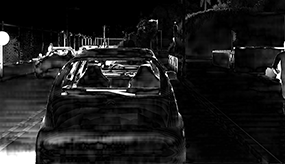}\\

\end{tabular}
\caption{Disparity prediction results on HR-VS. For each method, we show both the predicted disparity map (top) and the error map (bottom). For the error map, the darker the color, the lower the end point error (EPE).
%For error map, the pixel value is set to 255 (white) if the EPE is larger than 10, and the darker the color, the lower the EPE. 
}
\label{fig:disp_hrvs}
\end{figure*}

\begin{figure*}[t!]
\centering
\begin{tabular}{ccc}
\hspace{-3mm}Left image & \hspace{-2mm}PSMNet & \hspace{-2mm}Ours\_joint \\

% \hspace{-3mm}image/GT & SPS-Stereo & DispNetC & Ours \\
\hspace{-3mm}\includegraphics[height=1.25in]{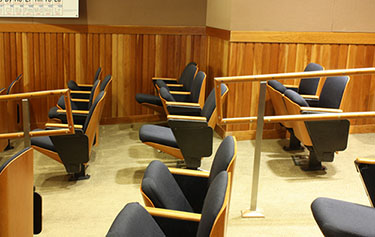}  &
\hspace{-2.8mm}\includegraphics[height=1.25in]{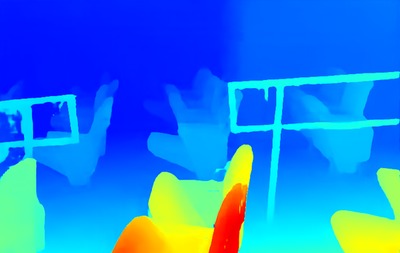} & 
\hspace{-2.8mm}\includegraphics[height=1.25in]{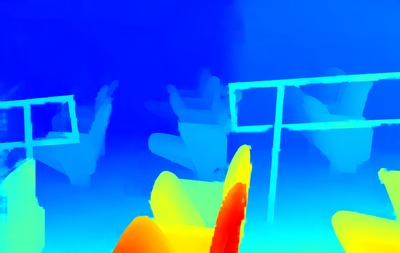} \\
& \hspace{-2.8mm}\includegraphics[height=1.25in]{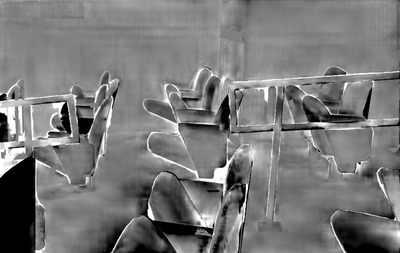} &
\hspace{-2.8mm}\includegraphics[height=1.25in]{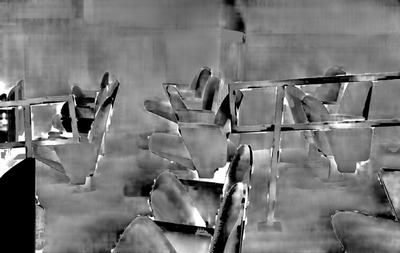} \\

\hspace{-3mm}\includegraphics[height=1.4in]{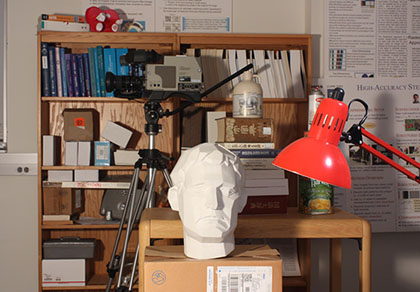}&
\hspace{-2mm}\includegraphics[height=1.4in]{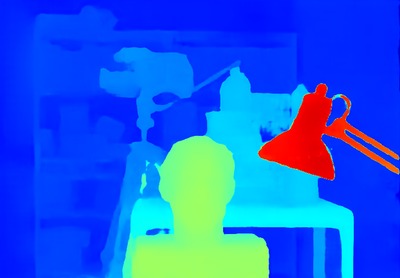} &
\hspace{-2mm}\includegraphics[height=1.4in]{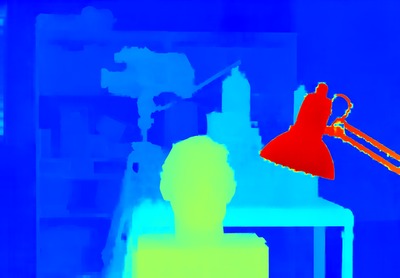} \\
& \hspace{-2mm}\includegraphics[height=1.4in]{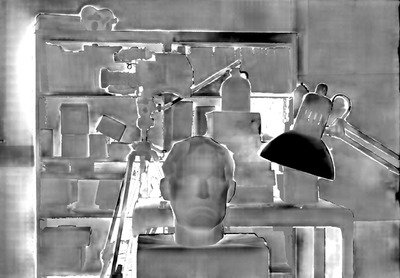} &
\hspace{-2mm}\includegraphics[height=1.4in]{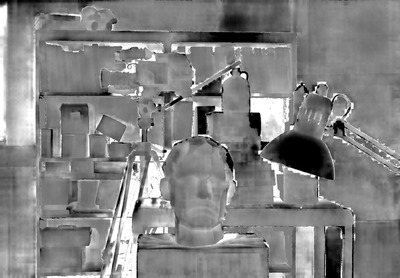} \\

\hspace{-3mm}\includegraphics[height=1.4in]{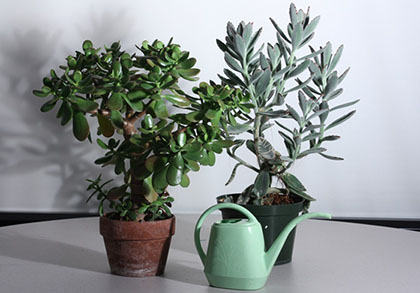} & 
\hspace{-2mm}\includegraphics[height=1.4in]{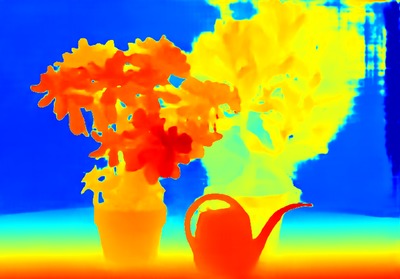} &
\hspace{-2mm}\includegraphics[height=1.4in]{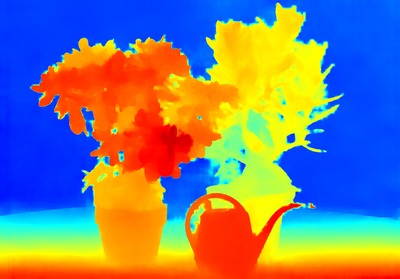}\\
& \hspace{-2mm}\includegraphics[height=1.4in]{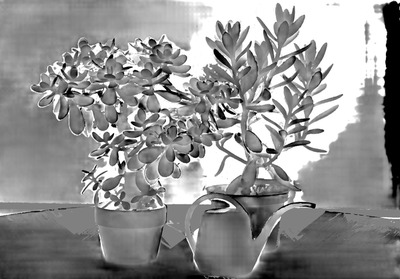} &
\hspace{-2mm}\includegraphics[height=1.4in]{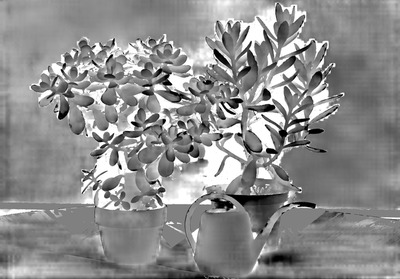} \\

\end{tabular}
\caption{Disparity estimation results on Middlebury-v3. For each method, we show both the predicted disparity map (top) and the error map (bottom). For the error map, the darker the color, the lower the error. All the images are from Middlebury-v3 leaderboard.
%For error map, the pixel value is set to 255 (white) if the EPE is larger than 10, and the darker the color, the lower the EPE. 
}
\label{fig:disp_mb}
\end{figure*}

Figure~\ref{fig:disp_sceneflow}, Figure~\ref{fig:disp_hrvs}, and Figure~\ref{fig:disp_mb} show the disparity prediction results on SceneFlow, HR-VS and Middlebury-v3, respectively. Compared to PSMNet, our methods are able to better preserve the fine details, such as the headset wire (the seventh row of Figure~\ref{fig:disp_sceneflow}) , street lamp post (the first row of Figure~\ref{fig:disp_hrvs}) and the leaves (the fifth row of Figure~\ref{fig:disp_mb}). We also observe that our method can better handle textureless areas, such as the car back in the seventh row of Figure~\ref{fig:disp_hrvs}. It is probably because our method directly downsample the images 16 times before sending them to the modified PSMNet, while the original PSMNet only downsamples the image 4 times, and uses stride-2 convolution to perform another $4\times$ downsampling later. The input receptive filed (w.r.t. the original image) of our method is actually larger than that of original PSMNet, which enables our method to better leverage context information around the textureless area. 

\begin{figure*}[t!]
\centering
\begin{tabular}{ccc}
\hspace{-3mm}Left image &  \hspace{-3mm}Ours\_fixed & \hspace{-3mm}Ours\_joint \\

% \hspace{-3mm}image/GT & SPS-Stereo & DispNetC & Ours \\
\hspace{-3mm}\includegraphics[width=2.2in]{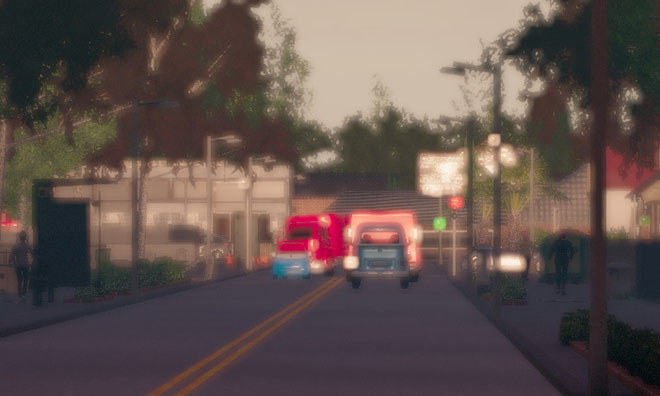}  &
\hspace{-3mm}\includegraphics[width=2.2in]{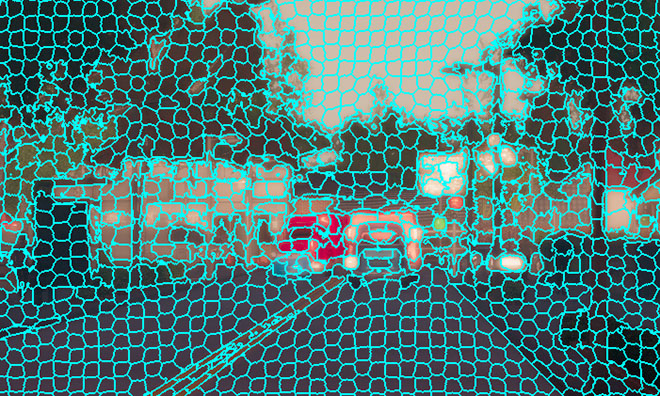} & 
\hspace{-3mm}\includegraphics[width=2.2in]{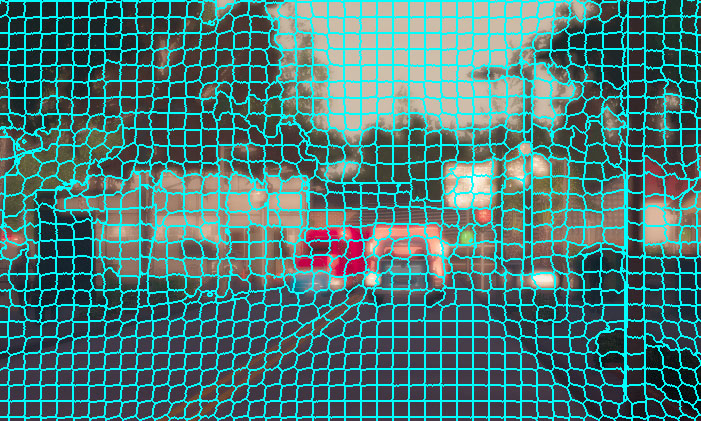} \\

 \hspace{-3mm}\includegraphics[width=2.2in]{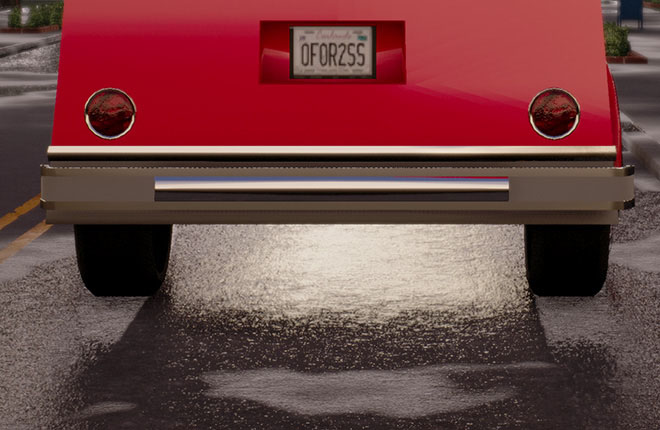} & 
\hspace{-3mm}\includegraphics[width=2.2in]{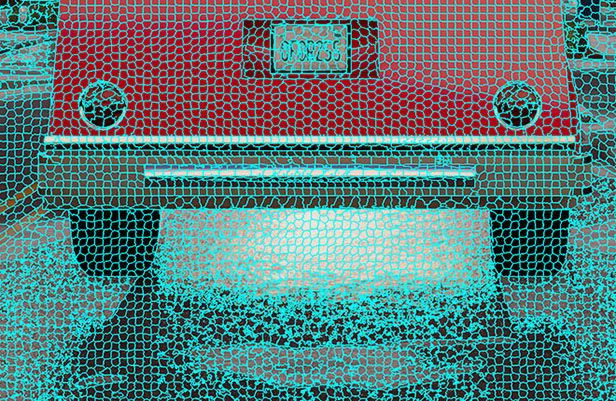} &
\hspace{-3mm}\includegraphics[width=2.2in]{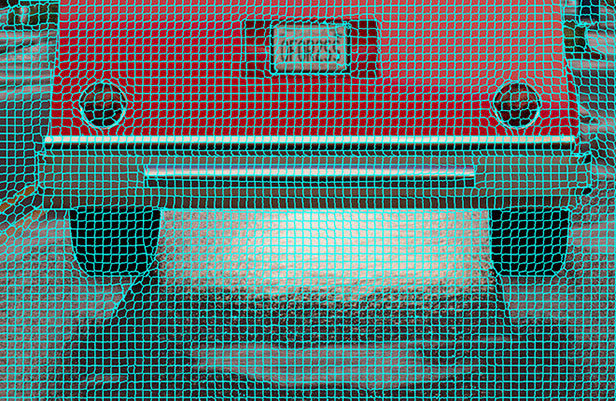} \\ 

\hspace{-3mm}\includegraphics[width=2.2in]{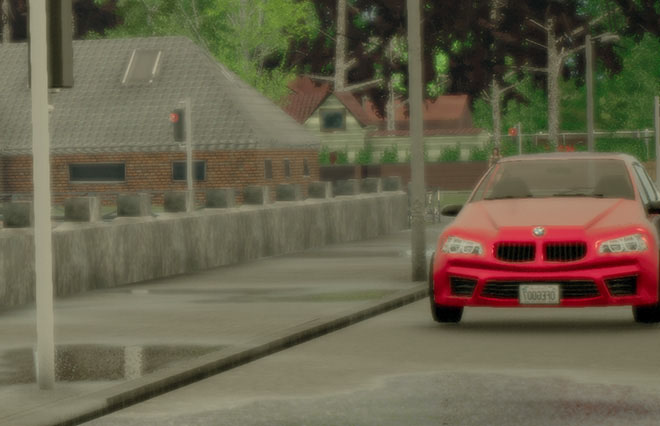} &
\hspace{-3mm}\includegraphics[width=2.2in]{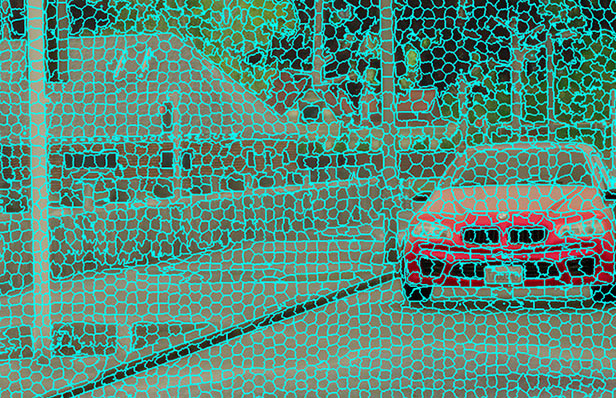} &
\hspace{-3mm}\includegraphics[width=2.2in]{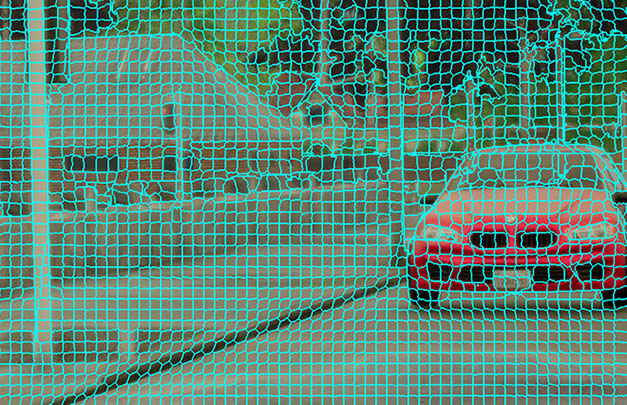} \\

\hspace{-3mm}\includegraphics[width=2.2in]{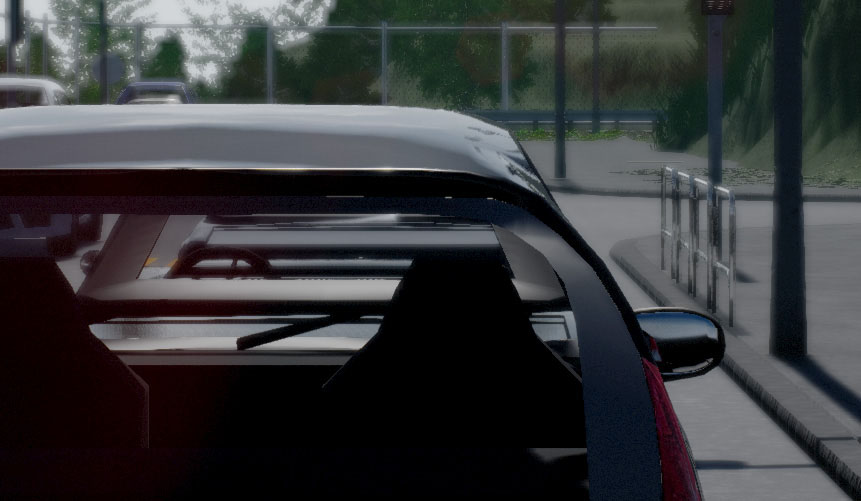} &
\hspace{-3mm}\includegraphics[width=2.2in]{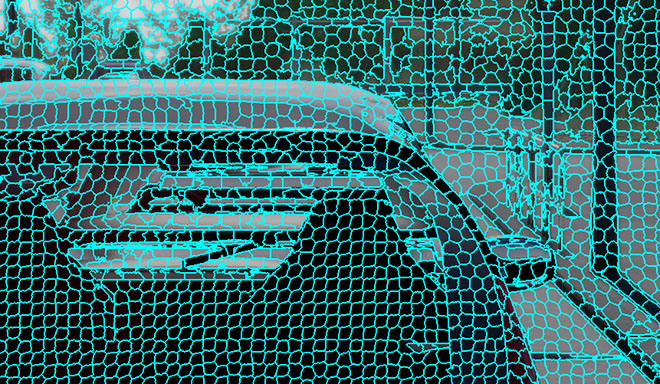} &
\hspace{-3mm}\includegraphics[width=2.2in]{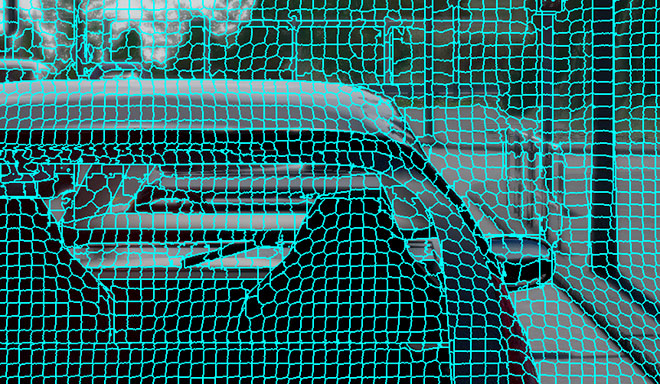}
\\

% Ours\_fixed & \hspace{-3mm}\includegraphics[height=1.2in, width=2in]{figures/supp_material/HRVS/fixed_spixel/exp-2_w-4_pos-79-14_00400_spixel.png} &
%  \\

% Ours\_joint & \hspace{-3mm}\includegraphics[height=1.2in]{figures/supp_material/HRVS/joint_spixel/exp-2_w-4_pos-79-14_00400_spixel.png} &
%  \\
% \arrayrulecolor{red} \hline \\ [-1.8ex]

\end{tabular}
\caption{Comparsion of superpixel segmentation results on HR-VS. Note we do not enforce the superpixel connectivity here.
% \textbf{Top rows}: images; \textbf{Middle rows}:  {Ours\_fixed} method; \textbf{Bottom rows}: {Ours\_joint} method.
%For error map, the pixel value is set to 255 (white) if the EPE is larger than 10, and the darker the color, the lower the EPE. 
}
\label{fig:spixel_hrvs}
\end{figure*}

Figure~\ref{fig:spixel_hrvs} visualizes the superpixel segmentation results of \textbf{Ours\_fixed} and \textbf{Ours\_joint} methods on HR-VS dataset. In general, Superpixels generated by \textbf{Ours\_joint} are more compact and pay more attentions to the disparity boundary. The color boundaries that are not aligned with the disparity boundary, such as the water pit on the road in the second row of Figure~\ref{fig:spixel_hrvs}, are often ignored by \textbf{Ours\_joint}. This phenomenon reflects the influence of disparity estimation on the superpixels in the joint training.

% Further, we note that the importance of superpixel segmentation to our method can also be observed by comparing the corresponding images in Figure~\ref{fig:disp_hrvs} and Figure~\ref{fig:spixel_hrvs}. Some details, such as the wall pier (the fifth row of Figure~\ref{fig:spixel_hrvs}) and the bicycle dock (the seventh row of Figure~\ref{fig:disp_hrvs}), are missing in our disparity prediction, as the superpixels generated by our methods fail to preserve the corresponding boundaries (the third and fourth row of Figure~\ref{fig:spixel_hrvs}).

%=========bsd==========

%=========nyu==========

%=========flyingthings==========

%=========  HRVS  ==========

%=========  HRVS spixel  ==========

%=========  Mb-v3 disp  ==========

\end{document}